\documentclass[preprint, 10pt]{elsarticle}

\usepackage[margin=2.5cm]{geometry}
\usepackage{setspace}
\usepackage{lmodern}
\usepackage{amsmath,amssymb}
\usepackage{amsfonts,amsthm}
\usepackage{lineno}
\usepackage{graphicx}
\usepackage{bm, bbm}
\usepackage{algorithm2e}
\usepackage{physics}
\usepackage{color}
\usepackage{pxfonts}
\usepackage[footnotesize,bf]{caption}
\usepackage{subcaption}
\usepackage{mathtools}
\usepackage{hhline}
\usepackage[T1]{fontenc}

\newcommand{\D}{D}
\newtheorem{theorem}{Theorem}[section]
\newcommand{\E}{\mathbb{E}}

\newcommand{\Real}{\textrm{I\!R}}
\newcommand{\overbar}[1]{\mkern 1.5mu\overline{\mkern-1.5mu#1\mkern-1.5mu}\mkern 1.5mu}

\RestyleAlgo{boxruled}
\graphicspath{{figures/}}
\journal{arXiv.org}

\begin{document}
\begin{frontmatter}
\title{Learning in Modal Space: Solving Time-Dependent Stochastic PDEs Using Physics-Informed Neural Networks}
\author[Brown]{Dongkun Zhang}
\author[SNH]{Ling Guo}
\author[Brown]{George Em Karniadakis \corref{cor}} \ead{george\_karniadakis@brown.edu}
\address[Brown]{Division of Applied Mathematics, Brown University, Providence RI, USA}
\address[SNH]{Department of Mathematics, Shanghai Normal University, Shanghai, China}
\cortext[cor]{Corresponding Author}

\begin{abstract}
One of the open problems in scientific computing is the long-time integration of nonlinear stochastic partial differential equations (SPDEs), especially with arbitrary initial data. We address this problem by taking advantage of recent advances in scientific machine learning and the spectral dynamically orthogonal (DO) and bi-orthogonal (BO) methods for representing stochastic processes. The recently introduced DO/BO methods reduce the SPDE into solving a system of deterministic PDEs and a system of stochastic ordinary differential equations. Specifically, we propose two new Physics-Informed Neural Networks (PINNs) for solving time-dependent SPDEs, namely the NN-DO/BO methods. The proposed methods incorporate the DO/BO constraints into the loss function (along with the modal decomposition of the SPDE) with an implicit form instead of generating explicit expressions for the temporal derivatives of the DO/BO modes. Hence, the NN-DO/BO methods can overcome some of the drawbacks of the original DO/BO methods. For example, we do not need the assumption that the covariance matrix of the random coefficients is invertible as in the original DO method, and we can remove the assumption of no eigenvalue crossing as in the original BO method. Moreover, the NN-DO/BO methods can be used to solve time-dependent stochastic inverse problems with the same formulation and same computational complexity as for forward problems. We demonstrate the capability of the proposed methods via several numerical examples, namely: (1) A linear stochastic advection equation with deterministic initial condition: we obtain good results with the proposed methods while the original DO/BO methods cannot be applied directly in this case. (2) Long-time integration of the stochastic Burgers' equation: we show the good performance of NN-DO/BO methods, especially the effectiveness of the NN-BO approach for such problems with many eigenvalue crossings during the whole time evolution, while the original BO method fails. (3) Nonlinear reaction diffusion equation: we consider both the forward problem and the inverse problems, including very noisy initial point values, to investigate the flexibility of the NN-DO/BO methods in handling inverse and mixed type problems. Taken together, these simulation results demonstrate that the NN-DO/BO methods can be employed to effectively quantify uncertainty propagation in a wide range of physical problems but future work should address the efficiency issue of PINNs for forward problems.
\end{abstract}

\begin{keyword}
scientific machine learning \sep data-driven modeling \sep dynamical orthogonality \sep bi-orthogonality \sep uncertainty quantification \sep inverse problems 
\end{keyword}

\end{frontmatter}

\section{Introduction}\label{S:1}
Physics-informed neural networks (PINNs)~\cite{Raissi2018} are a special class of PDE-induced networks that encode the physics (expressed by the PDE) into a deep neural network (DNN) that shares parameters with a standard DNN that approximates the quantity of interest (QoI), e.g. the solution of the PDE. In practice, this implies that the loss function that expresses mismatch in the labelled data is augmented by the residual of the PDE, which is represented efficiently by automatic differentiation and is evaluated at random points in the time-space domain. This approximation of the nonlinear operators by the DNN is justified theoretically based on the pioneering work of~\cite{ChenChen93, ChenChen95}, which goes well beyond the universal function approximation theorem of~\cite{Cybenko}. This simple and easy to program algorithm has been shown to be successful for diverse problems in physics and fluid mechanics~\cite{Raissi2018a, Raissi2017b, Raissi_JFM}, especially for inverse problems and even for discovering hidden physics~\cite{Hiddenfluidmechanics}. The advantages of encoding the PDE itself into a DNN are multiple: (1) we require much less data to train the DNN since we are searching for the minima on the manifold-solution of the PDE; (2) we respect the conservation laws of mass, momentum and energy; and most importantly, (3) we can truly predict the state of the system, unlike the DNNs driven solely by data that can interpolate accurately only within the training domain. While there is still a lot of work to be done to make PINNs efficient simulation machines, one of the main open issues is {\em uncertainty quantification} in predicting the QoI, which will reflect the various sources of uncertainty, i.e., from the approximation of the DNN to the data and physical model uncertainties.

In~\cite{dongkunNNapc2018} we addressed the issue of {\em total uncertainty} for first time and combined {\em dropout} and {\em arbitrary polynomial chaos} to model stochasticity in steady SPDEs. Here, we consider the more difficult case of time-dependent nonlinear SPDEs and hence we need to introduce a more effective way of dealing with the complexity of long-time integration of stochastic systems. To this end, we employ a generalized form of time-dependent Karhunen-lo\`{e}ve (KL) decomposition, first introduced in~\cite{SapsisPhd}, appropriate for second-order random fields, which has the form:
\begin{equation}\label{eqn:KL1}
    u(x,t;\omega) \approx \sum_{i=1}^{\infty}u_i(x,t) Y_{i}(t;\omega), \quad \omega \in \Omega.
\end{equation}
This approach can evolve the time-dependent basis of modes $u_i(x,t)$ and stochastic coefficients $Y_{i}(t;\omega)$ simultaneously, and it is different than the standard polynomial chaos methods~\cite{Xiu2002,Xiu2005,Xiu2010,narayan_stochastic_2015}. To remove the redundancy in this representation, we need some constraints. For example, this can be achieved by imposing \emph{dynamical constraints} on the spatial basis, which is the so-called ``dynamically orthogonal" (DO) methodology first proposed in \cite{Sapsis2009,Sapsis2012}. Alternatively, by imposing static constraints on both the spatial and stochastic basis, the ``bi-orthogonal" (BO) methodology was developed in~\cite{Cheng2013,Cheng2013a}. For both DO and BO we need to derive explicitly the evolution equations for all the components involved, i.e. the mean, spatial basis, and stochastic basis. The DO and BO formulations are mathematically equivalent~\cite{Choi2014}, but they exhibit computationally complimentary properties. Specifically, the BO formulation may fail due to crossing of the eigenvalues of the covariance matrix~\cite{Choi2014}, while both BO and DO become unstable when there is a high condition number of the covariance matrix or zero eigenvalues. A rigorous and sharp error bounds of DO method was first given by Zhou et al. in~\cite{musharbash2014dynamically}, where it was shown that the DO modes can capture the effective directions. For more applications and improvements of DO/BO methods, we refer to \cite{Sapsis2013jcp,Choi2014, Lermusiaux2016,Hessam2017} and references therein. 

The purpose of this paper is to combine PINNs and the DO/BO methodologies together to obtain new effective methods for solving time-dependent SPDEs -- we will refer to them as NN-DO/BO methods. Concretely, we first build a surrogate neural net for the solution of the time-dependent SPDEs based on the generalized KL expansion (Eq.~\ref{eqn:KL1}). Then the DO/BO constraints are included into the loss function and we train the neural network by minimizing this loss function to obtain the solution. Compared with the original DO/BO method, the merits of the proposed methods are the following:
\begin{itemize}
    \item We do not need the assumption in our NN-DO approach that the covariance matrix of the random coefficients is invertible, even for SPDEs with deterministic initial conditions.
    \item We can deal with eigenvalue crossing in the given time domain when applying our NN-BO approach.
    \item The same NN-DO/BO formulation and computer code can be applied for solving time-dependent stochastic {\em inverse} problems or problems driven by sparse noisy data, with the same computational complexity.
\end{itemize}

The organization of this paper is as follows. In Section~\ref{S:2}, we set up the time-dependent stochastic problems. In Section~\ref{S:3}, we give a brief review of the dynamically orthogonal and bi-orthogonal methodologies. In Section~\ref{S:4}, we formulate our NN-DO/BO framework after the introduction of the PINNs for solving deterministic differential equations. In Section~\ref{S:5}, we provide a detailed study of the accuracy and performance of the NN-DO/BO approach with numerical examples. We include two benchmark cases that are specifically designed to have exact solution for the DO and BO representations, followed by a nonlinear stochastic forward problem with high input stochastic dimensionality and noisy data as the initial condition, and a nonlinear inverse problem where we try to identify the model parameters. Finally, we conclude with a brief discussion in Section~\ref{S:6}.

\section{Problem Setup}\label{S:2}
Let $(\Omega,\mathcal{F},P)$ be a probability space, where $\Omega$ is the sample space, $\mathcal{F}$ is the $\sigma$-algebra of subsets of $\Omega$, and $P$ is a probability measure. Let $\D$ be a bounded domain in $\Real^{d}$ ($d = 1,2,$ or 3) whose boundary is denoted by $\partial \D$, and $[0,T]$ be the time domain of interest. We consider the following time-dependent SPDE:
\begin{equation}\label{eqn:sPDE}
    \pdv{u}{t} = \mathcal{N}_x[u(x,t;\omega)], \quad x \in \D,\, t \in [0, T],\, \omega \in \Omega,
\end{equation}
with initial and boundary conditions:
\begin{align}
u \left(x,t;\omega \right) &= u_0(x;\omega), & t &= t_0, \label{eqn:ic}\\
\mathcal{B}_x\left[u(x,t;\omega)\right] &= h(x,t;\omega), & x &\in \partial\D, \label{eqn:bc}
\end{align}
where $\mathcal{N}_x$ is a differential operator and $\mathcal{B}_x$ is a linear differential operator acting on the domain boundary. Assume that our quantity of interest, $u(x,t;\omega)$, is a second-order random field. The initial and boundary conditions for Eq.~\ref{eqn:sPDE} are denoted by $u_0(x;\omega)$ and $h(t,x;\omega)$. Our aim is to solve Eq.~\ref{eqn:sPDE}, and specifically, to evaluate the mean and standard deviation of the solution $u(x,t;\omega)$.

\section{An Overview of the DO and BO Decomposition Methods}\label{S:3}
For a random field $u(x,t;\omega)$ that evolves in time, the generalized Karhunen-Lo\`eve (KL) expansion at a given time $t$ is
\begin{equation}\label{eqn:kl}
    u(x,t;\omega) = \overbar{u}(x,t) + \sum_{i=1}^{\infty}\sqrt{\lambda_i} \phi_i(x,t) \xi_{i}(t;\omega), \quad \omega \in \Omega,
\end{equation}
where $\overbar{u}$ is the mean, $\xi_{i}(t;\omega)$ ($i=1,2,3,1...$) are zero-mean independent random variables, $\lambda_i$ and $\phi_i$ are the $i^\text{th}$ largest eigenvalue and the corresponding eigenfunction of the covariance kernel, i,e., they solve the following eigenproblem:
\begin{equation}
    \int_{\D} C_{u(x_1,t)u(x_2,t)}\phi_i(x_2,t) \dd x_2 = \lambda_i \phi_i(x_1, t).
\end{equation}
Here $C_{u(x_1,t)u(x_2,t)} = \E \left[ (u(x_1,t;\omega)-\overbar{u}(x_1,t)) (u(x_2,t;\omega)-\overbar{u}(x_2,t)) \right]$ is the covariance kernel of $u$.

Next we consider a generalized expansion first proposed in~\cite{Sapsis2009}:
\begin{equation}\label{eqn:fullexpansion}
u(x,t;\omega) = \overbar{u}(x,t) + \sum_{i=1}^{\infty}u_i(x,t) Y_{i}(t;\omega), \quad \omega \in \Omega.
\end{equation}
Similar to the KL expansion, the random field $u(x,t;\omega)$ is decomposed into two parts: (i) the deterministic mean field function $\overbar{u}(x,t)$, and (ii) the random fluctuation part consists of an infinite summation of deterministic orthogonal fields $u_i(x,t)$ with 0-mean stochastic coefficients $Y_i(t;\omega)$. Formally, we have
\begin{equation}\label{eqn:mean}
\overbar{u}(x,t) = \E[u(x,t;\omega)] = \int_{\Omega}u(x,t;\omega) \dd{P(\omega)},
\end{equation}
\begin{equation}\label{eqn:ortho}
\langle u_i, u_j \rangle = 0, \quad \text{for}\ i \ne j\ \text{and}\ i,j=1,2,\dots,
\end{equation}
and
\begin{equation}\label{eqn:center}
\E[Y_i] = 0, \quad \text{for}\ i = 1,2,\dots .
\end{equation}

We define the linear subspace $V_{S} = \operatorname{span} \left\{ u_{i}(x,t) \right\}_{i=1}^{N}$ as the linear space spanned by the first $N$ deterministic bases. For now let us assume that $Y_i(t;\omega)$ are linearly independent and $\Omega_{S} = \operatorname{span} \left\{ Y_i(t;\omega) \right\}_{i=1}^{N}$ is the linear subspace in $L^2(\Omega)$ spanned by the first $N$ stochastic coefficients. The truncated expansion $u_N(x,t;\omega)$, defined by
\begin{equation}\label{eqn:expansion}
    u_N(x,t;\omega) = \overbar{u}(x,t) + \sum_{i=1}^{N}u_i(x,t)Y_{i}(t;\omega),
\end{equation}
is the projection of $u(x,t;\omega)$ to the subspace $V_S\times\Omega_S$. Without making any assumptions on their form, the governing equations Eq.~\ref{eqn:sPDE} and Eq.~\ref{eqn:mean}--\ref{eqn:center} represent the only information that can be utilized to derive the evolution equations of $u_i$ and $Y_i$. Note that both the stochastic coefficients $Y_{i}(t;\omega)$ and the orthogonal bases $u_{i}(x,t)$ are time-dependent (and they are evolving according to the system dynamics), unlike the standard polynomial chaos where the stochastic coefficients are time-independent. There exists some redundancy in the Eq.~\ref{eqn:expansion}, and therefore, additional constraints need to be imposed in order to formulate a well posed problem for the unknown quantities. Here we review the DO and BO approaches, which have different assumptions on the constraints.

\subsection{Dynamically Orthogonal (DO) Representation}
As first proposed in~\cite{Sapsis2009}, a natural constraint to overcome redundancy is that the evolution of the bases $\left\{u_{i}(x,t)\right\}_{i=1}^{N}$ be orthogonal to the space $V_{S}$; this can be expressed through the following dynamically orthogonal (DO) condition:
\begin{equation}\label{eqn:DOcondition}
\frac{dV_{S}}{dt}\perp V_{S} \Longleftrightarrow \left\langle \frac{\partial u_{i}(x,t)}{\partial t}, u_{j}(x, t) \right\rangle = 0 \quad i,j = 1,\dots,N.
\end{equation}
Here $\left\langle u(x, t), v(x ,t) \right\rangle$ is defined as the spatial inner product $\int_D u(x, t)v(x ,t)\dd{x}$. Comparing Eq.~\ref{eqn:expansion} with the standard KL expansion Eq.~\ref{eqn:kl}, in the DO representation, we set $u_i$ to have unit length and $Y_i$ carries the scaling coefficient as the result of the eigenvalues. Note that the DO condition preserves the orthonormality and the length of the bases $\left\{u_{i}(x,t)\right\}_{i=1}^{N}$ since
\begin{equation}
\pdv{t} \left\langle u_{i}(\cdot, t), u_{j}(\cdot ,t) \right\rangle = \left\langle \pdv{u_i(\cdot ,t)}{t}, u_{j}(\cdot ,t) \right\rangle + \left\langle u_{i}(\cdot,t), \pdv{u_j(\cdot, t)}{t} \right\rangle = 0, \quad i,j=1,\dots,N.
\end{equation}
It is proved in~\cite{Sapsis2009} that the DO condition leads to a set of independent and explicit evolution equations for all the unknown quantities. Here we state the DO evolution equations without proof:
\begin{theorem}[see~\cite{Sapsis2009}]
Under the assumptions of the DO representation, the original SPDE (Eq.~\ref{eqn:sPDE}) is reduced to the following system of equations:
\begin{equation}\label{eqn:do}
\begin{aligned}
&\pdv{\overbar{u}(t,x)}{t} = \E[\mathcal{N}_x[u(\cdot,t;\omega)],\\
&\dv{Y_{i}(t;\omega)}{t} = \langle\mathcal{N}_x[u(\cdot,t;\omega)]-\E[\mathcal{N}_x[u(\cdot,t;\omega)], u_{i}(\cdot,t)\rangle, &i=1,\dots,N,\\
&\sum_{i=1}^{N}C_{Y_{i}(t)Y_{j}(t)}\pdv{u_i(t,x)}{t} = \prod_{V_{s}^{\bot}} \E \left[ \mathcal{N}_x \left[u(\cdot,t;\omega)\right] Y_{j}\right], &j=1,\dots,N.
\end{aligned}
\end{equation}
\end{theorem}
The projection in the orthogonal complement of the linear subspace $V_S$ is defined as
$\prod_{V_{\mathrm{S}}^{\perp}}F(x) = F(x)-\prod_{V_{s}}F(x)=F(x)-\sum_{k=1}^{N}\left\langle F (\cdot), u_{k}(\cdot,t)\right\rangle u_{k}(\cdot,t),$ and the covariance of the stochastic coefficients is $C_{Y_{i}(t)Y_{j}(t)}=E\left[Y_{i}(t; \omega)Y_{j}(t; \omega)\right]$. The associated boundary conditions are determined by
\begin{equation}
\begin{gathered}
\mathcal{B}_x\left[\overbar{u}(x,t;\omega)\right]|_{x\in \partial \D} = \E[h(x,t; \omega)],\\
\mathcal{B}_x\left[u_{i}(x, t)\right] |_{x\in\partial \D} = \E\left[Y_{j}(t; \omega) h(x,t; \omega) \right]C_{Y_{i}(t)Y_{j}(t)}^{-1}, \quad \text{for $i=1,2,\dots,N$,}
\end{gathered}
\end{equation}
and the initial conditions at $t=t_0$ for the DO components are given by
\begin{equation}\label{eqn:doinit}
\begin{aligned}
&\overbar{u}\left(x, t_{0}\right)= \E\left[u_{0}(x; \omega) \right],\\
&Y_{i}\left(t_{0}; \omega\right)= \langle u_{0}(\cdot, \omega)-\overbar{u}\left(x, t_{0}\right), v_{i}(\cdot) \rangle,\\
&u_{i}\left(x, t_{0} \right)=v_{i}(x),
\end{aligned}
\end{equation}
for all $i = 1,\dots, N$, where $v_{i}(x)$ is the eigenfields of the standard KL expansion of $u(x,t_0;\omega)$.

It is shown in~\cite{Sapsis2009} that by imposing suitable restrictions on the DO representation, the equations for methods such as Polynomial Chaos (PC) or Proper Orthogonal Decomposition (POD) can be recovered from the DO evolution equations. For example, PC can be recovered by setting $Y_i(t;\omega) = \Psi_i(\xi(\omega))$, where $\Psi_i(\xi)$ is an orthogonal polynomial in terms of $\xi$. Moreover, it is shown in~\cite{ChoiThesis} that there exists an one-to-one correspondence between the eigenvalues of the KL expansion for $u(x,t;\omega)$ and the eigenvalues of the covariance matrix $C_{Y_i(t)Y_j(t)}$ in the DO representation given any fixed time $t$. Thus, the stochastic coefficients $Y_i$ together with the modes $u_i$ provide the necessary information to describe both the shape and the magnitude of the uncertainty that characterizes a stochastic field but also the principle directions in which the stochasticity is distributed.

The moments of $u(x,t;\omega)$ can be readily computed from the DO representation. For example, the first moment, i.e., the mean, can be trivially obtained from the first term $\overbar{u}(x,t)$, and the variance can be calculated as follows:
\begin{equation}
    \operatorname{Var}[u] = \E\left[(u-\overbar{u})^2\right] = \E\left[\left(\sum_{i=1}^N u_i Y_i\right)^2\right] = \sum_{i,j=1}^N u_i \E[Y_i Y_j]u_j.
\end{equation}

\subsection{Bi-Orthogonal (BO) Representation}
An alternative way to overcome the aforementioned redundancy in Eq.~\ref{eqn:expansion} is the Bi-Orthogonal (BO) condition, which imposes the static constraint that both the spatial basis functions and stochastic coefficients are orthogonal in time~\cite{Cheng2013a}. In other words, we have the following conditions:
\begin{equation}\label{eqn:BOcondition}
\langle u_i(\cdot,t), u_j(\cdot,t)\rangle =\lambda_i(t)\delta_{ij},\quad \E[Y_iY_j]=\delta_{ij}, \quad i,j=1,\dots,N,
\end{equation}
where the $\lambda_i$s are eigenvalues of the covariance kernel and $\delta_{ij}$ is the Dirac's delta function. There is a slight difference between the DO and BO representation: the stochastic coefficients carry the eigenvalues of the covariance operator in the DO representation while the spatial bases carry the eigenvalues of the covariance operator in the BO representation.

Next, we define the matrix $S$ and $M$ whose entries are
\begin{equation}\label{eqn:S_M}
    S_{ij} = \left\langle u_i, \pdv{u_j}{t} \right\rangle, \quad M_{ij} = \E\left[ Y_i \dv{Y_j}{t} \right].
\end{equation}
Then by taking the time derivative of Eq.~\ref{eqn:BOcondition}, we have
\begin{equation}\label{eqn:bo}
    \begin{aligned}
    &S_{ij} + S_{ji} = \left\langle u_i, \pdv{u_j}{t} \right\rangle + \left\langle \pdv{u_i}{t}, u_j \right\rangle = 0,\quad &\text{for}\ i\ne j,\\
    &S_{ij} = \frac{1}{2}\dv{\lambda_i(t)}{t}, &\text{for}\ i = j,\\
    &M_{ij} + M_{ji} = \E\left[ Y_i \dv{Y_j}{t} \right] + \E\left[ \dv{Y_i}{t} Y_j \right] = 0.
    \end{aligned}
\end{equation}
Here we state the BO evolution equations without proof:
\begin{theorem}[see~\cite{Cheng2013}]
We assume that the bases and stochastic coefficients satisfy the BO condition. Then, the original SPDE (Eq.~\ref{eqn:sPDE}) is reduced to the following system of equations:
\begin{equation}
\begin{aligned}
&\pdv{\overbar{u}(t,x)}{t} = \E[\mathcal{N}_x[u(\cdot,t;\omega)]],\\
&\lambda_i \dv{Y_{i}(t;\omega)}{t} = -\sum_{j=1}^{N}S_{ij}Y_j + \left\langle \mathcal{N}_x[u] - \E \left[ \mathcal{N}_x[u] \right], u_i(\cdot,t) \right\rangle, &i=1,\dots,N,\\
&\pdv{u_i(t,x)}{t} = -\sum_{j=1}^{N}M_{ij}u_j + \E \left[ \mathcal{N}_x[u] Y_i \right], &i=1,\dots,N.
\end{aligned}
\end{equation}
Moreover, if $\lambda_i \ne \lambda_j$ for $i \ne j,\ i,j,=1,2,\dots,N$, the $N$-by-$N$ matrices $S$ and $M$ have closed form expression:
\begin{equation}
    \begin{aligned}
    &M_{ij} = \begin{cases}
        \frac{G_{ij} + G_{ji}}{-\lambda_i + \lambda_j}, \qquad & \text{if $i \ne j$}\\
        0, & \text{if $i = j$}
    \end{cases},\\
    &S_{ij} = \begin{cases}
        G_{ij} + \lambda_i M_{ij}, & \text{if $i \ne j$}\\
        G_{ii}, & \text{if $i = j$}
    \end{cases},
    \end{aligned}
\end{equation}
where the matrix $G_{ij}$ is defined as $\left\langle\E\left[\mathcal{N}_x[u]Y_j\right], u_i\right\rangle$.
\end{theorem}
Similar to the DO method, the boundary condition is given by
\begin{equation}
\begin{gathered}
\mathcal{B}_x\left[\overbar{u}(x,t;\omega)\right]|_{x\in \partial \D} = \E[h(x,t; \omega)],\\
\mathcal{B}_x\left[u_{i}(x, t)\right] |_{x\in\partial \D} = \E\left[Y_{i}(t; \omega) h(x,t; \omega) \right], \quad \text{for $i=1,2,\dots,N$,}
\end{gathered}
\end{equation}
and the initial condition is generated from the KL expansion of $u_0(x;\omega)$.

\subsection{A Brief Summary of DO/BO Methods}
Let us note the following difference between the DO and BO condition: the spatial bases, $u_i$, under the DO condition evolve to the direction which is normal to the space $V_S$ they expand (orthogonality is automatically maintained), while under the BO condition, only the mutual orthogonality within $u_i$ and within $Y_i$ are required and there is no restriction to the direction of their evolution. As a compensation to the lack of constraints in spatial bases, the BO condition puts an additional orthogonality restriction on the random coefficients $Y_i$. However, Choi et al.~\cite{Choi2014} have proved theoretically the equivalence between the DO and the BO methods, in a sense that one method is an exact reformulation of the other via a differential transformation.

Each method can be applied to a limited range of problems since the evolution equations are valid only if certain assumptions are satisfied. For the DO method, it is assumed that the covariance matrix of random coefficients, $C_{Y_iY_j}$, is invertible. Therefore, it fails when applied to some benchmark problems such as a stochastic PDE with deterministic initial condition. This is because all coefficients $Y_i$ are equal to 0 at the initial state and their covariance matrix is singular. For the BO method, it is assumed that there is no eigenvalue crossing in the given time domain in order to calculate the explicit expression of $M_{ij}$ and $S_{ij}$. However, some strategies have been proposed to get around those issues, e.g., the hybrid gPC-DO method~\cite{Choi2013} and the psedo-inverse hybrid BO-DO method~\cite{Hessam2017} are developed to address the above limitations.

Inspired by the DO and BO methods, we introduce a new procedure for solving time-dependent stochastic PDEs within the framework of Physics-Informed Neural Networks (PINNs). The proposed methods inherit the similarities between the DO and the BO methods and can be implemented with either the DO or the BO version that are free from the aforementioned restrictions.

\section{Methodology}\label{S:4}
\subsection{Physics-informed neural network}
In this part, we briefly review using DNNs to solve the deterministic differential equations~\cite{Lagaris1998, Lagaris2000, Raissi2017a}, and its generalization for solving deterministic inverse problems in~\cite{Raissi2017b}. Suppose that we have a parameterized deterministic differential equation:
\begin{equation}\label{eqn:ODE}
\begin{gathered}
    \mathcal{N}[u;\eta]=0,\quad x\in \D,\\
    \text{B.C.:}\qquad \mathcal{B}_x[u] = 0, \quad x\in\Gamma,
\end{gathered}
\end{equation}
where $u(x)$ is the solution and $\eta$ denotes the parameters.

\begin{figure}[htbp]
  \centering
  \includegraphics[width=0.6\linewidth]{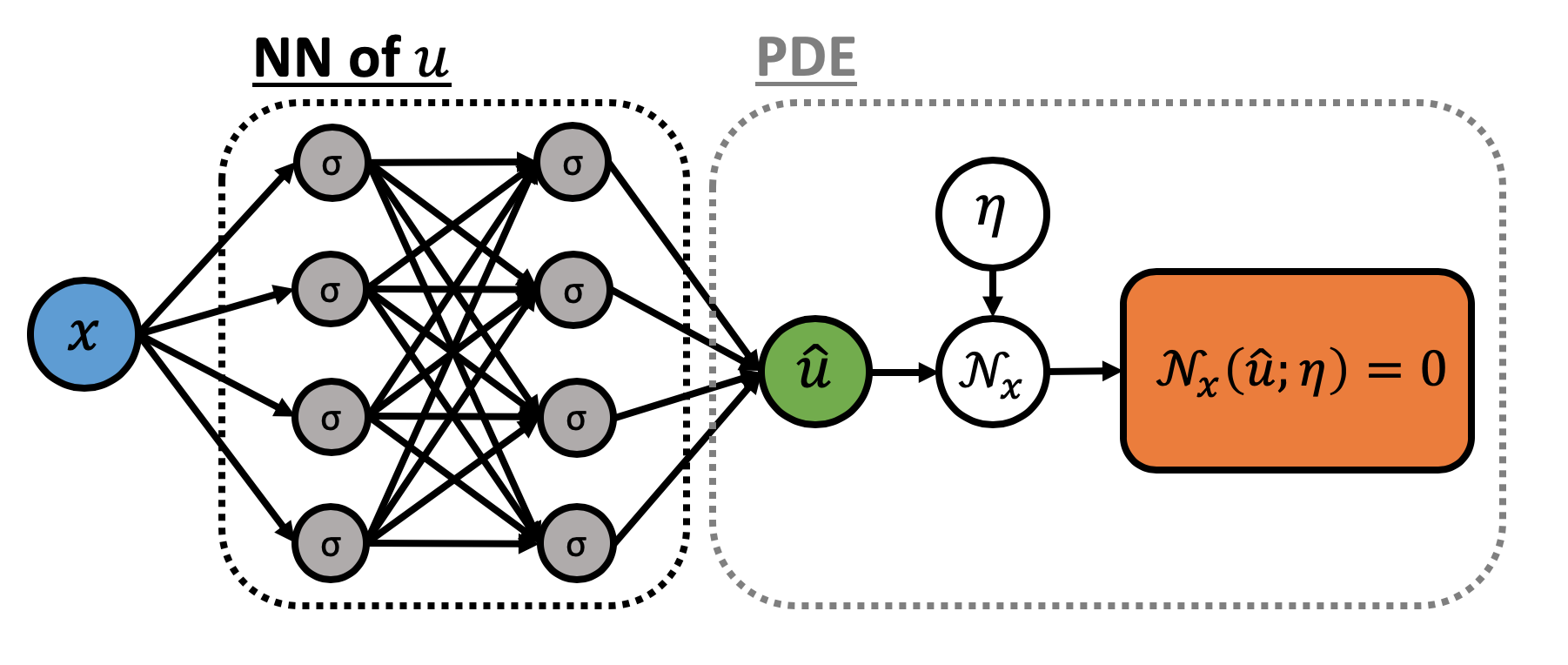}
  \caption{Schematic of the PINN for solving differential equations.}\label{fig:pinn_sketch}
\end{figure}

A DNN, denoted by $\hat{u}(x;\theta)$, is constructed as a surrogate of the solution $u(x)$, and it takes the coordinate $x$ as the input and outputs a vector that has the same dimension as $u$. Here we use $\theta$ to denote the DNN parameters that will be tuned at the training stage, namely, $\theta$ contains all the weights $\bm{w}$ and biases $\bm{b}$ in $\hat{u}(x;\theta)$. For this surrogate network $\hat{u}$, we can take its derivatives with respect to its input by applying the chain rule for differentiating compositions of functions using the automatic differentiation, which is conveniently integrated in many machine learning packages such as Tensorflow~\cite{Abadi2016}. The restrictions on $\hat{u}$ is two-fold: first, given the set of scattered data of the $u(x)$ observations, the network should be able to reproduce the observed value, when taking the associated $x$ as input; second, $\hat{u}$ should comply with the physics imposed by Eq.~\ref{eqn:ODE}. The second part is achieved by defining a residual network:
\begin{equation}\label{eqn:PINN}
    \hat{f}(x;\theta,\eta):= \mathcal{N}[\hat{u}(x;\theta);\eta],
\end{equation}
which is computed from $\hat{u}$ straightforwardly with automatic differentiation. This residual network $\hat{f}$, also named the physics-informed neural network (PINN), shares the same parameters $\theta$ with network $\hat{u}$ and should output the constant 0 for any input $x\in\D$. Figure~\ref{fig:pinn_sketch} shows a sketch of the PINN. At the training stage, the shared parameters $\theta$ (and also $\eta$, if it is also to be inferred) are fine-tuned to minimize a loss function that reflects the above two constraints.

Suppose we have a total number of $N_u$ observations on $u$, collected at location $\{x_u^{(i)}\}_{i=1}^{N_u}$, and $N_c$ is the number of training points $\{x_f^{(i)}\}_{i=1}^{N_c}$ where we evaluate the residual $\hat{f}(x_f^{(i)}; \theta, \eta)$. We shall use $(x^*, y^*)$ to represent a single instance of training data, where the first entry $x^*$ denotes the input and the second entry $y^*$ denotes the anticipated output (also called ``label''). The workflow of solving a differential equation with PINN can be summarized as follows:

\begin{algorithm}[H]
\textbf{Step 1:} Specify the training set:
  \begin{equation*}
    \hat{u}\text{ network: }\{(x_u^{(i)}, u(x_u^{(i)}))\}_{i=1}^{N_u}, \quad \hat{f}\text{ network: }\{(x_f^{(i)}, 0)\}_{i=1}^{N_f};
  \end{equation*}\\
\textbf{Step 2:} Construct a DNN $\hat{u}(x;\theta)$ with random initialized parameters $\theta$;\\
\textbf{Step 3:} Construct the residual network $\hat{f}(x;\theta,\eta)$ by substituting the surrogate $\hat{u}$ into the governing equation (Eq.~\ref{eqn:PINN}) via automatic differentiation and arithmetic operations;\\
\textbf{Step 4:} Specify a loss function by summing the mean squared error of both the $u$ observations and the residual:
  \begin{equation}\label{eqn:PINN_loss}
    \mathcal{LOSS}(\theta,\eta)=\frac{1}{N_u}\sum_{i=1}^{N_u}[\hat{u}(x_u^{(i)};\theta)-u(x_u^{(i)})]^2+\frac{1}{N_c}\sum_{i=1}^{N_c}\hat{f}(x_f^{(i)};\theta,\eta)^2;
  \end{equation}\\
\textbf{Step 5:} Train the DNN to find the best parameters $\theta$ and $\eta$ by minimizing the loss function:
  \begin{equation}
    \theta=\arg\min \mathcal{LOSS}(\theta,\eta).
  \end{equation}
\caption{PINN for solving deterministic PDEs}
\end{algorithm}

\subsection{A weak formulation interpretation of the DO and BO methods}\label{S:3-1}
The derivation of equations for both the DO and the BO methods can be summarized into four steps as follows:
\begin{enumerate}
    \item Apply operator $\E[\cdot]$ on both sides of the SPDE and replace $u$ by the finite expansion in Eq.~\ref{eqn:expansion}. Notice that $\E[Y_i]=0$. This leads to the first equation in Eq.~\ref{eqn:do} and Eq.~\ref{eqn:bo}:
    \begin{equation}\label{eqn:weakmean}
        \E \left[ \pdv{u}{t} \right] = \pdv{\overbar{u}}{t} = \E \left[ \mathcal{N}_x[u(x,t;\omega)] \right]
    \end{equation}
    \item Apply operator $\langle \cdot, u_i \rangle$ on both sides of the SPDE:
    \begin{equation}\label{eqn:weakui}
        \left\langle \pdv{u}{t}, u_i \right\rangle = \left\langle \mathcal{N}_x[u(x,t;\omega)], u_i \right\rangle
    \end{equation}
    \item Apply operator $\E[\cdot Y_i]$ on both sides of the SPDE:
    \begin{equation}\label{eqn:weakyi}
        \E \left[ \pdv{u}{t} Y_i\right] = \E \left[ \mathcal{N}_x[u(x,t;\omega)] Y_i \right]
    \end{equation}
    \item Substitute $u$ by the truncated expansion in Eq.~\ref{eqn:expansion}, and use the DO and BO constraints (Eq.~\ref{eqn:DOcondition} and Eq.~\ref{eqn:BOcondition}) to simplify Eq.~\ref{eqn:weakui} and Eq.~\ref{eqn:weakyi}.
\end{enumerate}
Due to the orthogonality of $u_i(x,t)$, they form a valid set of basis in the physical space $\D$. The random coefficients $Y_i(t;\omega)$ are also linearly independent as they are orthogonal under the BO representation, and the DO representation is equivalent to the BO representation so $Y_i$ will not degenerate in the DO expansion either. Therefore, the random coefficients $Y_i(t;\omega)$ form a valid set of basis in the probability space $L^2(\Omega)$. Consequently, Eq.~\ref{eqn:weakmean}--\ref{eqn:weakyi} are the weak formulation of the original SPDE in the physical space and the probability space (note that Eq.~\ref{eqn:weakmean} is the inner product of both side of the original SPDE on constant $1$, which can be regarded as the $0^\text{th}$ basis in the probability space), and they provide all the necessary information to find the solution $u_N$ in $V_S\times\Omega_S$.

\subsection{NN-DO/BO Methods}\label{S:3-2}
In this section we formalize the algorithm of solving time-dependent stochastic PDEs using PINNs. First, we rewrite Eq.~\ref{eqn:expansion} as
\begin{equation}\label{eqn:expan}
    u_N(x,t;\omega) = \overbar{u}(x,t) + \sum_{i=1}^{N}a_i(t)u_i(x,t)Y_{i}(t;\omega),
\end{equation}
while enforcing $\langle u_i, u_i \rangle = 1$ and $\E[Y_i^2] = 1$. The time-dependent coefficients $a_i(t)$ are scaling factors and play the role of $\sqrt{\lambda_i}$ when we compare Eq.~\ref{eqn:expan} with the standard KL expansion Eq.~\ref{eqn:kl}. Suppose that the original SPDE is parameterized into a PDE that involves a finite set of random variables $\xi(\omega)$, then $u(x,t;\omega)$ can be written as $u(x,t;\xi)$ and $Y_i(t;\omega)$ can be written as $Y_i(t;\xi)$. Four separate neural networks are constructed:
\begin{enumerate}
    \item The neural net $\overbar{u}_{nn}(x, t)$ that takes $x$ and $t$ as the input and outputs $\E[u(x,t;\omega)]$;
    \item The neural net $A_{nn}(t)$ that takes $t$ as the input and outputs a $N$-dimensional vector representing $a_i(t)$, for $i=1,2,\dots,N$;
    \item The neural net $U_{nn}(x, t)$ that takes $x$ and $t$ as the input and outputs a $N$-dimensional vector representing $u_i(x,t)$, for $i=1,2,\dots,N$;
    \item The neural net $Y_{nn}(\xi, t)$ that takes $\xi$ and $t$ as the input and outputs a $N$-dimensional vector representing $Y_i(t;\xi)$, for $i=1,2,\dots,N$.
\end{enumerate}
A surrogate neural net for the solution $u_N(x,t;\omega)$ can be constructed from those four neural nets by substituting them into Eq.~\ref{eqn:expan}, yielding
\begin{equation}\label{eqn:reconstruction}
    u_{nn}(x,t;\xi) = \overbar{u}_{nn} + \sum_{i=1}^N A_{nn,i} U_{nn,i} Y_{nn,i}.
\end{equation}
Since the weak formulation of SPDE involves integration in both the physical and the probability spaces, the neural nets are evaluated at the physical training points $\{x_c^k\}_{k=1}^{n_x}$ and the probabilistic training points $\{\xi_c^l\}_{l=1}^{n_{\xi}}$, where $n_x$ and $n_{\xi}$ are the numbers of training points. In the time domain $[0, T]$ we uniformly sample $n_t$ random points $\{t_c^s\}_{s=1}^{n_t}$. Once we have constructed the computation graph, the derivatives of the quantity of interest with respect to time $t$ and space coordinate $x$ can be easily obtained via the auto-differentiation algorithm, and the integration terms can be evaluated by using a numerical quadrature rule.

The loss function is a weighted summation of four components: the weak formulation of SPDE, initial/boundary conditions, constraints on $U_{nn}$ and $Y_{nn}$, and the additional regularization terms. The loss function in each part consists of mean squared errors (MSEs) associated with the prescribed constraints, calculated from the sampled training points. Next, we will illustrate each of these four components of loss function and write down their explicit expressions.

\subsubsection{Loss Function for the Weak Formulation of SPDE}
The weak form of the SPDE, i.e., Eq.~\ref{eqn:weakmean}--Eq.~\ref{eqn:weakyi} can be rewritten as
\begin{equation}
    \epsilon_1^{ks} \coloneqq \E \left[ \pdv{u_{nn}}{t} (x_c^k,t_c^s;\xi) - \mathcal{N}_x[u_{nn}(x_c^k,t_c^s;\xi)] \right] = 0,
\end{equation}
\begin{equation}
    \epsilon_2^{sl} \coloneqq\left\langle \pdv{u_{nn}}{t} (x,t_c^s;\xi_c^l) - \mathcal{N}_x[u_{nn}(x,t_c^s;\xi_c^l)], U_{nn,i}(x,t_c^s) \right\rangle = 0,
\end{equation}
\begin{equation}
    \epsilon_3^{ks} \coloneqq\E \left[ \left(\pdv{u_{nn}}{t} (x_c^k,t_c^s;\xi) - \mathcal{N}_x[u_{nn}(x_c^k,t_c^s;\xi)] \right) Y_{nn,i}(t_c^s;\xi) \right] = 0,
\end{equation}
where the integration in the physical space and the probability space shall be evaluated by using a numerical quadrature rule. The first part of the loss function is calculated by
\begin{equation}\label{eqn:weakloss}
    \text{MSE}_{\text{w}} = \frac{1}{n_x n_t} \sum_{k, s} \left(\epsilon_1^{ks}\right)^2 + \frac{1}{n_t n_{\xi}} \sum_{s, l} \left(\epsilon_2^{sl}\right)^2 + \frac{1}{n_x n_t} \sum_{k, s} \left(\epsilon_3^{ks}\right)^2.
\end{equation}

\subsubsection{Loss Function for Initial and Boundary Conditions}
Let $t_0$ be the initial time of computation. The initial condition for the representation in Eq.~\ref{eqn:expan} is similar to Eq.~\ref{eqn:doinit}. The only difference is that $Y_i$ are normalized to have unit variance, and the standard deviation of $\langle u_0(x;\xi)-\overbar{u}(x,t_0), v_i(x) \rangle$ is assigned to be the initial value for $a_i$. Here $v_i(x)$ are the normalized KL modes for $u(x,t_0;\omega)$, and they are the initial value for $u_i$. That is,
\begin{equation}
    \begin{aligned}
    &\overbar{u}(x,t_0) = \E\left[u(x, t_0; \xi)\right],\\
    &u_i(x,t_0) = v_i(x),\\
    &a_i(t_0) = \sqrt{\E\left[ \left\langle u(x, t_0; \xi) - \E[u(x, t_0; \xi)], v_i \right\rangle ^2 \right]},\\
    &Y_i(t_0;\xi) = \frac{1}{a_i(t_0)} \left\langle u(x, t_0; \xi) - \E[u(x, t_0; \xi)], v_i \right\rangle.
    \end{aligned}
\end{equation}
For deterministic initial condition, $u_i(x, t_0)$ are set to be orthonormal bases satisfying the boundary condition, $Y_i(t_0; \xi)$ are set to be the gPC bases of $\xi$ with unit variance, and $a_i(t_0)$ is set to be $0$. The initial condition shall be imposed to the neural network by adding an extra penalty term $\text{MSE}_{\text{IC}}$, and it is calculated as follows:
\begin{equation}
\begin{aligned}
    \text{MSE}_{\text{IC}} = &\frac{1}{n_x}\sum_{k=1}^{n_x}\left(\overbar{u}_{nn}(x_c^k, t_0) - \overbar{u}(x_c^k, t_0)\right)^2 + \frac{1}{Nn_x}\sum_{i=1}^N\sum_{k=1}^{n_x}\left(U_{nn,i}(x_c^k, t_0) - u_i(x_c^k, t_0)\right)^2 \\
    &+\frac{1}{N}\sum_{i=1}^N \left(A_{nn,i}(t_0) - a_i(t_0) \right)^2 + \frac{1}{Nn_{\xi}} \sum_{i=1}^N\sum_{l=1}^{n_{\xi}} \left(Y_{nn,i}(t_0;\xi_c^l) - Y_i(t_0;\xi_c^l)\right)^2.
\end{aligned}
\end{equation}

The boundary condition is imposed by taking the weak formulation of Eq.~\ref{eqn:bc} in the random space, i.e.,
\begin{equation}
\begin{aligned}
    \E\left[\mathcal{B}_x[u(x_b,t;\xi)]\right] &= \E[h(x_b, t;\xi)] \Rightarrow \mathcal{B}_x\left[\overbar{u}(x_b, t)\right] = \E[h(x_b, t;\xi)],\\
    \E\left[\mathcal{B}_x[u(x_b, t;\xi)]Y_i(t;\xi)\right] &= \E[h(x_b, t;\xi)Y_i(t;\xi)]\\
    &\Rightarrow \sum_{j=1}^N C_{Y_iY_j}a_j(t)\mathcal{B}_x\left[u_j(x_b, t)\right] = \E[h(x_b, t;\xi)Y_i(t;\xi)].
\end{aligned}
\end{equation}
Thus, the loss associated with the boundary condition is
\begin{equation}
\begin{aligned}
    \text{MSE}_{\text{BC}} = &\frac{1}{n_t}\sum_{s=1}^{n_t}\left(\mathcal{B}_x[\overbar{u}_{nn}(x_b,t_c^s)]-\E\left[h(x_b,t;\xi)\right]\right)^2\\
    &+ \frac{1}{Nn_t}\sum_{i=1}^N\sum_{s=1}^{n_t}\left(\sum_{j=1}^N C_{Y_iY_j}(t_c^s)A_{nn,j}(t_c^s)\mathcal{B}_x\left[U_{nn,j}(x_b, t_c^s)\right]-\E\left[h(x_b, t_c^s;\xi)Y_{nn,i}(t_c^s;\xi)\right]\right)^2,
\end{aligned}
\end{equation}
where the expectations and covariance matrix shall be evaluated by using a numerical quadrature rule.

Note that the periodic boundary condition can be strictly imposed by modifying the neural nets $\overbar{u}_{nn}$ and $U_{nn}$ by replacing the input $x$ with the combination of $\sin(2\pi x/L)$ and $\cos(2\pi x/L)$, where $L$ is the length of domain $\D$. This is because any continuous $2\pi$-periodic function can be written as a nonlinear function of $\sin(x)$ and $\cos(x)$. This modification simplifies the loss function by removing the loss due to the periodic boundary condition.

\subsubsection{Loss Function for the Constraints on $U_{nn}$ and $Y_{nn}$}
This is the part where we can have different implementations in favor of the DO or the BO method. Both DO and BO representations require that $\E[Y_i]=0$, and thus the loss functions in both implementations should involve the term
$\frac{1}{n_t}\sum_{s=1}^{n_t} \left(\E[Y_i(t_c^s;\xi)]\right)^2.$
For the DO constraint, Eq.~\ref{eqn:DOcondition} should be satisfied. In addition, we require that
\begin{equation*}
    \E\left[Y_i(t;\xi)\dv{Y_i(t;\xi)}{t}\right] = 0, \quad \text{$\forall t$ and $i=1,2,\dots,N$,}
\end{equation*}
so that $Y_i$ stay normalized with unit variance. The loss function for DO is:
\begin{equation}\label{eqn:do_loss}
\begin{aligned}
    \text{MSE}_{\text{DO}} = &\frac{1}{Nn_t}\sum_{i=1}^N\sum_{s=1}^{n_t} \left(\E[Y_i(t_c^s;\xi)]\right)^2\\
    &+ \frac{1}{N^2n_t}\sum_{i=1}^N\sum_{j=1}^N\sum_{s=1}^{n_t}\left\langle \dv{U_{nn,i}(x,t_c^s)}{t}, U_{nn,j}(x,t_c^s) \right\rangle^2\\
    &+ \frac{1}{Nn_t}\sum_{i=1}^N\sum_{s=1}^{n_t}\E\left[Y_{nn,i}(t_c^s;\xi)\dv{Y_{nn,i}(t_c^s;\xi)}{t}\right]^2.
\end{aligned}
\end{equation}
For the BO constraints, Eq.~\ref{eqn:bo} generates the following loss function:
\begin{equation}\label{eqn:bo_loss}
\begin{aligned}
    \text{MSE}_{\text{BO}} = &\frac{1}{Nn_t}\sum_{i=1}^N\sum_{s=1}^{n_t} \left(\E[Y_i(t_c^s;\xi)]\right)^2\\
    &+ \frac{1}{N^2n_t}\sum_{i=1}^N\sum_{j=1}^N\sum_{s=1}^{n_t}\left(\left\langle \dv{U_{nn,i}(x,t_c^s)}{t}, U_{nn,j}(x,t_c^s) \right\rangle + \left\langle \dv{U_{nn,j}(x,t_c^s)}{t}, U_{nn,i}(x,t_c^s) \right\rangle\right)^2\\
    &+ \frac{1}{Nn_t}\sum_{i=1}^N\sum_{j=1}^N\sum_{s=1}^{n_t}\left(\E\left[Y_{nn,i}(t_c^s;\xi)\dv{Y_{nn,j}(t_c^s;\xi)}{t}\right] + \E\left[Y_{nn,j}(t_c^s;\xi)\dv{Y_{nn,i}(t_c^s;\xi)}{t}\right]\right)^2.
\end{aligned}
\end{equation}
Since we put all the scaling factor to $a(t)$ and keep $u_i(x,t)$ normalized, $S_{ij} + S_{ji} = 0$ in Eq.~\ref{eqn:bo} still holds true when $i$ is equal to $j$.

\subsubsection{Loss Function for Additional Regularization}
Additional regularization terms shall be added to the loss function to reduce the risk of overfitting. Here we remark that it is helpful to add a penalty term from the original equation (Eq.~\ref{eqn:sPDE}) to speed up the training. The loss from the original equation is:
\begin{equation}
    \text{MSE}_0 = \frac{1}{n_xn_tn_{\xi}}\sum_{k=1}^{n_x}\sum_{s=1}^{n_t}\sum_{l=1}^{n_{\xi}}\left(\pdv{u_{nn}}{t}(x_c^k, t_c^s;\xi_c^l) - \mathcal{N}_x\left[u_{nn}(x_c^k, t_c^s;\xi_c^l)\right]\right)^2.
\end{equation}

\subsubsection{Putting the Loss Functions Together}
\begin{figure}[htbp]
  \centering
  \includegraphics[width=0.8\linewidth]{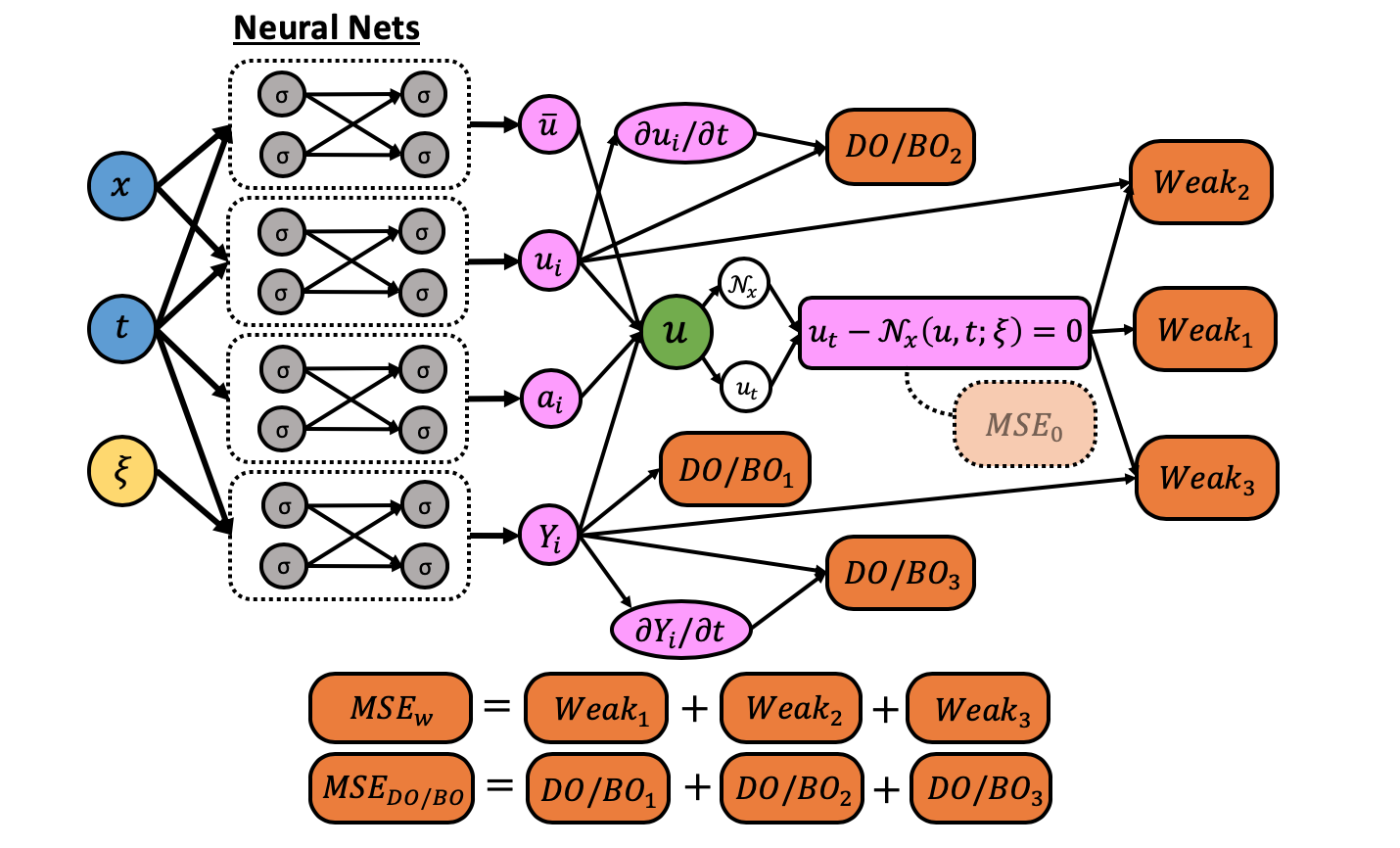}
  \caption{Schematic of the NN-DO/BO for solving time-dependent stochastic differential equations, where the blocks $\text{Weak}_1$, $\text{Weak}_2$, and $\text{Weak}_3$ correspond to the three right-hand side terms in Eq.~\ref{eqn:weakloss}, and blocks $\text{DO/BO}_1$, $\text{DO/BO}_2$, and $\text{DO/BO}_3$ correspond to the three right-hand side terms in Eq.~\ref{eqn:do_loss}/Eq.~\ref{eqn:bo_loss}, respectively.}\label{fig:sketch}
\end{figure}
A sketch of the computation graph for the loss functions $\text{MSE}_\text{w}$ and $\text{MSE}_{\text{DO/BO}}$ is shown in Figure~\ref{fig:sketch}. The loss function used for training the PINNs is the weighted summation of the aforementioned MSEs. Intuitively, we want to train the networks to gradually satisfy the weak formulation of the SPDE, while not violating the physical restrictions and the DO/BO constraints, as these are the cornerstones of the weak formulation. Therefore, we put a relatively large weight in front of $\text{MSE}_\text{IC}$, $\text{MSE}_\text{BC}$ and $\text{MSE}_\text{DO/BO}$, making their scale the same as that of $\text{MSE}_\text{w}$, if not slightly larger. The idea behind this is to remove the redundancy of Eq.~\ref{eqn:expan} in the first place. We put a small weight for the regularization term since it is only used to help speedup the training process, and is not essential. Nevertheless, the distribution of weights is still an open question for future research. In the numerical tests we train our neural nets by minimizing the following loss function:
\begin{equation}\label{eqn:loss}
    \mathcal{LOSS} = \text{MSE}_\text{w} + 100\times(\text{MSE}_\text{IC} + \text{MSE}_\text{BC} + \text{MSE}_\text{DO/BO}) + 0.1\times\text{MSE}_0.
\end{equation}

This proposed algorithm can be implemented with the DO or the BO constraints, and we name them the NN-DO or NN-BO method, respectively. The proposed algorithm is summarized as follows:

\begin{algorithm}[H]
\textbf{Step 1:} Build the neural networks for $\overbar{u}_{nn}(x,t)$, $A_{nn}(t)$, $U_{nn}(x,t)$ and $Y_{nn}(t;\xi)$;\\
\textbf{Step 2:} Select $n_x$ training points in the physical domain $\D$, $n_{\xi}$ training points in the stochastic space $\Omega$. Randomly pick $n_t$ points in the time domain $[0,T]$ from a uniform distribution;\\
\textbf{Step 3:} Specify the method to use (DO or BO) and calculate the loss function in Eq.~\ref{eqn:loss};\\
\textbf{Step 4:} Train the neural networks by minimizing the loss function;\\
\textbf{Step 5:} Reconstruct the SPDE solution using Eq.~\ref{eqn:reconstruction}.
\caption{NN-DO/BO for solving time-dependent stochastic PDEs}\label{alg:NN-DOBO}
\end{algorithm}

We remark that the bottleneck of the original DO/BO method is to generate an explicit expression for the temporal derivatives of the bases (Step 4 in Section~\ref{S:3-1}). For the standard DO method, it involves calculating the inverse of a covariance matrix which could be singular, and for the standard BO method, to obtain explicit expression for matrices $S$ and $M$ (Eq.~\ref{eqn:S_M}), one has to assume no eigenvalue crossing. In the proposed NN-DO/BO algorithm, there is no need to derive explicit expressions from constraints, instead we only need to write the constraints into the loss functions as they are.

\section{Simulation Results}\label{S:5}
We first test our NN-DO/BO methods with two benchmark cases that are especially designed to have exact solutions for the DO and BO representations. To demonstrate the advantage of the NN-DO/BO methods over the standard methods, we then solve a nonlinear diffusion-reaction equation with a 19-dimensional random input, where the problem is solved with very rough initial conditions given as discrete point values. Finally, an inverse problem is also considered to demonstrate the new capacity of the proposed NN-DO/BO methods. For all test cases we use deep feed-forward neural networks for $\overbar{u}_{nn}(x,t)$, $A_{nn}(t)$, $U_{nn}(x,t)$ and $Y_{nn}(t;\xi)$. The loss functions are defined in Eq.~\ref{eqn:loss}, and the Adam optimizer with learning rate 0.001 is used to train the networks.

\subsection{Application to a Linear Stochastic Problem}
In this section we present a pedagogical example by solving the linear stochastic advection equation using the NN-DO/BO methods. The stochastic advection equation with a random advection coefficient has the form
\begin{equation}\label{eqn:advection}
\begin{aligned}
&\pdv{u(x,t;\xi)}{t} + \xi \pdv{u(x,t;\xi)}{x} = 0, \quad \forall(x, t) \in \D\times[0, T],\\
&u(x,0;\xi)= -\sin (x), \quad \forall x \in \D,
\end{aligned}
\end{equation}
where the physical domain $\D$ is $[-\pi, \pi]$ and we obtain the solution until final time $T = \pi$. Periodic boundary conditions are considered, such that $u(-\pi, t) = u(\pi, t)$, $\forall t \in [0,T]$. The randomness comes from the advection velocity, which is modeled as a Gaussian random variable $\xi \sim N(0, \sigma^2)$ where we set $\sigma$ to be 0.8.

The exact solutions for the mean and variance of the stochastic advection equation, Eq.~\ref{eqn:advection}, can be calculated, and the closed form formulas of the DO and BO expansion components $u_i$ and $Y_i$, $i=1,2,\dots,N$ can be derived~\cite{ChoiThesis}. Here we write down the exact solution and the expansion components without giving details of the derivation:
\begin{itemize}
    \item Exact solutions:
    \begin{equation}\label{eqn:adv_exact}
        \begin{aligned}
        u(x,t;\xi) &= -\sin(x-\xi t)\\
        \E[u](x,t) &= -\sin(x)\exp\left(-\frac{\sigma^2 t^2}{2}\right)\\
        \operatorname{Var}[u](x,t) &= \frac{1}{2}\left[1-\cos(2x)\exp\left(-2\sigma^2 t^2\right)\right] - \E[u]^2
        \end{aligned}
    \end{equation}
    \item DO components:
    \begin{equation}
        u(x,t;\xi) = \E[u](x,t) + u_1^{DO}(x,t)Y_1^{DO}(t;\xi) + u_2^{DO}(x,t)Y_2^{DO}(t;\xi),
    \end{equation}
    where
    \begin{equation}\label{eqn:adv_do_exact}
        \begin{aligned}
        &u_1^{DO}(x,t) = -\frac{1}{\sqrt{\pi}}\cos(x), &&u_2^{DO}(x,t) = -\frac{1}{\sqrt{\pi}}\sin(x),\\
        &Y_1^{DO}(t;\xi) = -\sqrt{\pi}\sin(\xi t), &&Y_2^{DO}(t;\xi) = \sqrt{\pi}\left(\cos(\xi t) - \exp\left( -\frac{\sigma^2t^2}{2} \right)\right).
        \end{aligned}
    \end{equation}
    \item BO components:
    \begin{equation}
        u(x,t;\xi) = \E[u](x,t) + u_1^{BO}(x,t)Y_1^{BO}(t;\xi) + u_2^{BO}(x,t)Y_2^{BO}(t;\xi),
    \end{equation}
    where
    \begin{equation}\label{eqn:adv_bo_exact}
        \begin{aligned}
        &u_1^{BO}(x,t) = -\frac{\alpha_1(t)}{\sqrt{\pi}}\cos(x), &&u_2^{BO}(x,t) = -\frac{\alpha_2(t)}{\sqrt{\pi}}\sin(x),\\
        &Y_1^{BO}(t;\xi) = -\frac{\sqrt{\pi}}{\alpha_1(t)}\sin(\xi t), &&Y_2^{BO}(t;\xi) = \frac{\sqrt{\pi}}{\alpha_2(t)}\left(\cos(\xi t)-\exp\left(-\frac{\sigma^2t^2}{2}\right)\right),
        \end{aligned}
    \end{equation}
    and the normalizing factors
    \begin{equation*}
    \alpha_1(t) = \sqrt{\pi\E\left[sin^2(\xi t)\right]},\quad \alpha_2(t) = \sqrt{\pi\E\left[\left(\cos^2(\xi t)-\exp\left(-\frac{\sigma^2t^2}{2}\right)\right)^2\right]}.
    \end{equation*}
\end{itemize}

We set $n_t$, $n_x$ and $n_\xi$ all to be 50. The data points in the time domain $\{t_c^s\}_{s=1}^{n_t}$ are sampled from a uniform distribution. The training points $\{x_c^k\}_{k=1}^{n_x}$ are equidistantly distributed in $[-\pi, \pi]$. For the training points in the stochastic space, instead of using the Gauss-Hermite quadrature rule, we generate $\{\xi_c^l\}_{l=1}^{n_{\xi}}$ by applying the inverse cumulative distribution function of the standard normal distribution to the Gauss-Legendre quadrature points in $[0, 1]$, because the generated $\xi$ will be more concentrated near the origin, making it easier to train the neural networks. The neural networks are trained with an Adam optimizer (learning rate 0.001) for 300000 epochs.

\subsubsection{Case 1: NN-DO Method}
The standard DO method cannot be directly applied to this SPDE with deterministic initial condition. However, by applying the NN-DO method we obtain good results. Considering Eq.~\ref{eqn:expan}, the initial conditions are
\begin{equation}
    \begin{aligned}
    &\overbar{u}(x,0) = u(x,0),\, &&a_1(0) = a_2(0) = 0,\\
    &u_1(x, 0) = -\frac{1}{\pi}\cos(x), &&u_2(0) = -\frac{1}{\pi}\sin(x),\\
    &Y_1(0;\xi) = -\xi, &&Y_2(0;\xi) = -\frac{\sqrt{2}}{2}(\xi^2-1),
    \end{aligned}
\end{equation}
where we use the periodic orthonormal bases in the $[-\pi, \pi]$ interval as the initial conditions for $u_1$ and $u_2$, and we use the normalized Hermite polynomials for the initial conditions of $Y_1$ and $Y_2$. The neural networks $\overbar{u}_{nn}(x,t)$ and $U_{nn}(x,t)$ have three hidden layers with 32 neurons per hidden layer, the network $A_{nn}(t)$ has three hidden layers with 16 neurons per hidden layer, and the network $Y_{nn}(t;\xi)$ has four hidden layers with 64 neurons in each hidden layer. The reference solutions for the mean, variance, and the modes $u_i$ are taken directly from Eq.~\ref{eqn:adv_exact} and Eq.~\ref{eqn:adv_do_exact}. The reference values for the normalizing factors, $a_i$, are the standard deviations of $Y_i^{DO}$ in Eq.~\ref{eqn:adv_do_exact}, and the reference values for $Y_i$ are calculated by $Y_i^{DO}/a_i$.

\begin{figure}[htbp]
    \centering
    \begin{subfigure}{.5\textwidth}
        \centering
        \includegraphics[width=\linewidth]{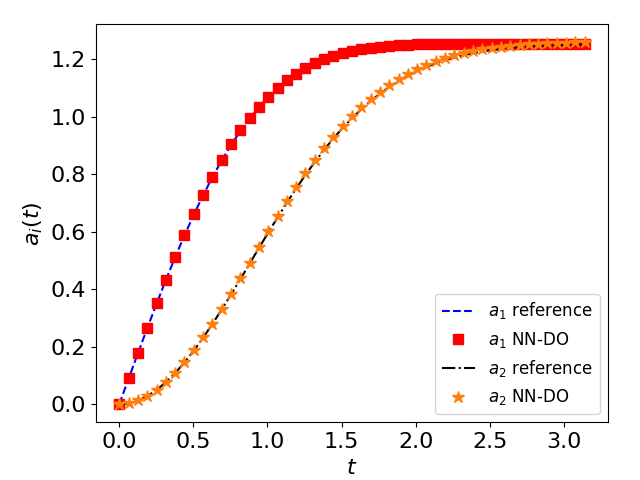}
        \caption{}\label{fig:adv_do_ai}
    \end{subfigure}%
    \begin{subfigure}{.5\textwidth}
        \centering
        \includegraphics[width=\linewidth]{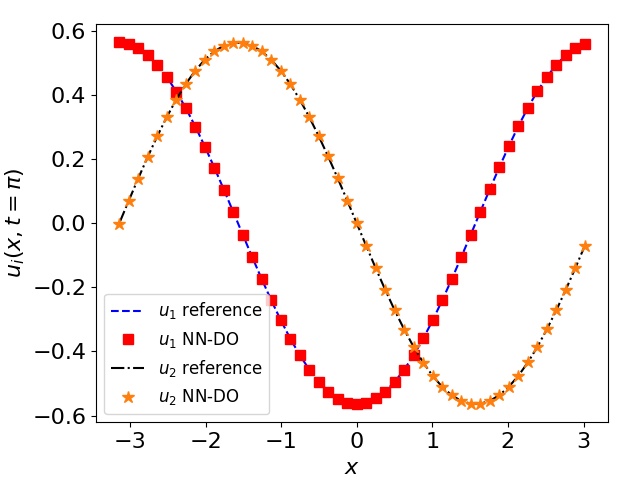}
        \caption{}\label{fig:adv_do_ui}
    \end{subfigure}
    \caption{Stochastic advection equation (NN-DO). Left: Plot of the evolution of the scaling factors $a_i$. They start from zero because of the deterministic initial condition, and increase with time, indicating that the randomness in the SPDE solution accumulates as time grows; Right: Plot of the bases $u_i$ at the final time $T=\pi$ versus the exact solutions. The scattered points for $u_i$ denote the training points in the physical space.}
\end{figure}

\begin{figure}[htbp]
    \centering
    \begin{subfigure}{.5\textwidth}
        \centering
        \includegraphics[width=\linewidth]{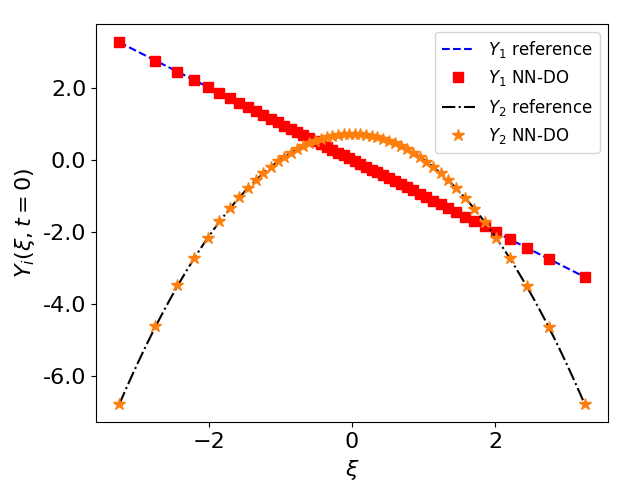}
        \caption{}
    \end{subfigure}%
    \begin{subfigure}{.5\textwidth}
        \centering
        \includegraphics[width=\linewidth]{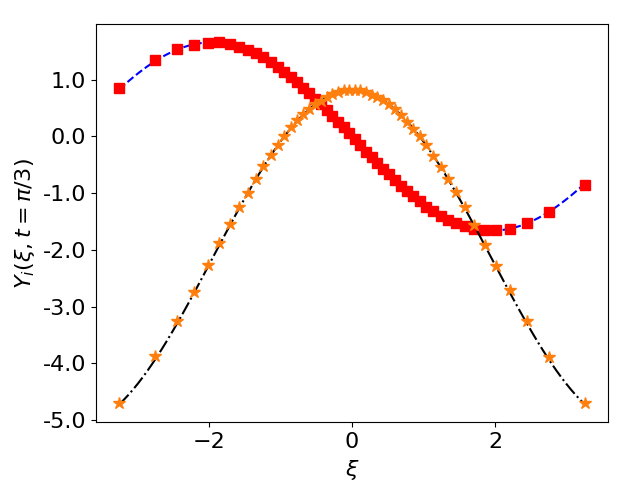}
        \caption{}
    \end{subfigure}
    \begin{subfigure}{.5\textwidth}
        \centering
        \includegraphics[width=\linewidth]{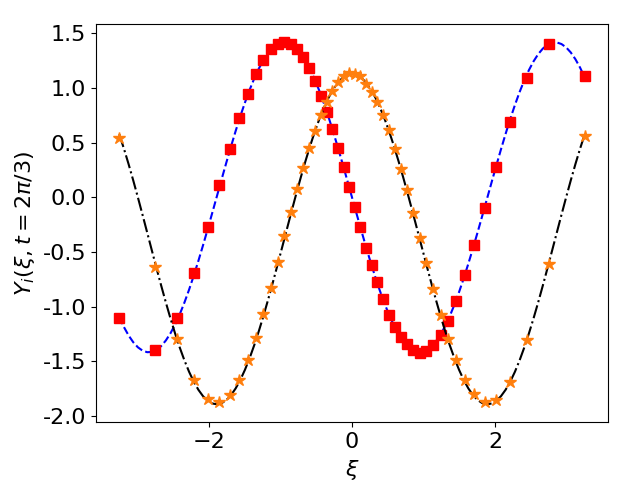}
        \caption{}
    \end{subfigure}%
    \begin{subfigure}{.5\textwidth}
        \centering
        \includegraphics[width=\linewidth]{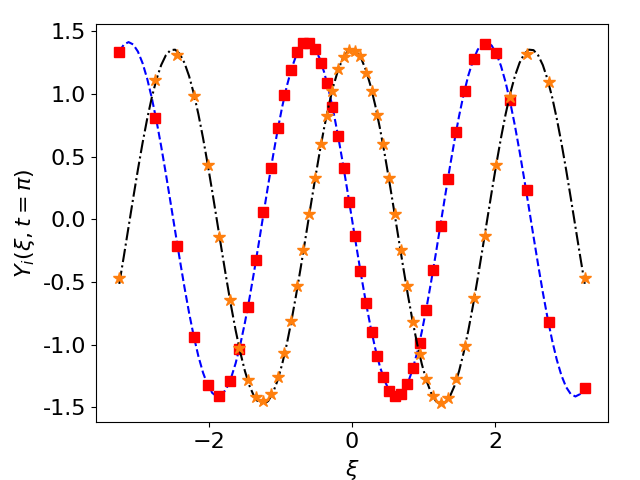}
        \caption{}
    \end{subfigure}
    \caption{Stochastic advection equation (NN-DO). Solutions for the random coefficients $Y_1$ and $Y_2$ at four different times $t=0, \pi/3, 2\pi/3 \text{ and } \pi$. Both of them agree with the exact solution. The scattered points denote the training points in the probabilistic space.}\label{fig:adv_do_yi}
\end{figure}

\begin{figure}[htbp]
    \centering
    \begin{subfigure}{.5\textwidth}
        \centering
        \includegraphics[width=\linewidth]{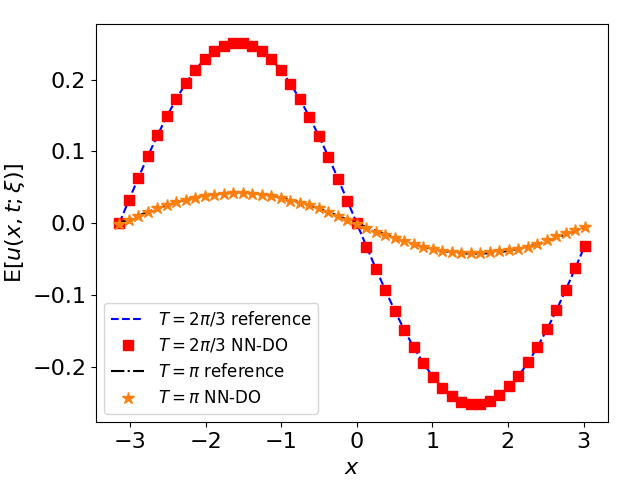}
        \caption{}\label{fig:adv_do_mean}
    \end{subfigure}%
    \begin{subfigure}{.5\textwidth}
        \centering
        \includegraphics[width=\linewidth]{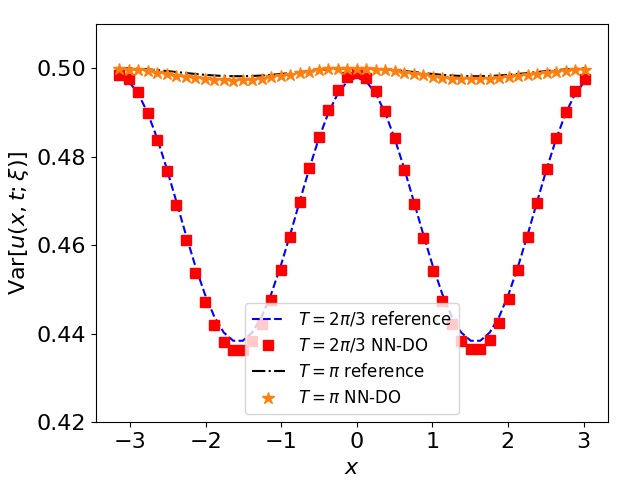}
        \caption{}\label{fig:adv_do_var}
    \end{subfigure}
    \caption{Stochastic advection equation (NN-DO). Mean and variance of the solution at time $T=2\pi/3$ and $T=\pi$. The scattered points denote the training points in the physical space.}\label{fig:adv_do_mean_var}
\end{figure}

We compare the results obtained from the NN-DO method with the exact solutions. Figure~\ref{fig:adv_do_ai} shows the evolution of the scaling factors $a_i$ ($i=1,2$) with time; they increase monotonically and converge at $T=\pi$, indicating that the randomness in the system grows from zero to fully developed state during the time period $t\in[0, \pi]$, as a result of the stochastic advection coefficient. Figure~\ref{fig:adv_do_ui} shows the comparison of the DO bases obtained from the NN-DO method and the exact bases at $T=\pi$. The DO bases generated by the neural networks agree well with the reference solutions. Figure~\ref{fig:adv_do_yi} shows the comparison of the stochastic coefficients $Y_i$ ($i=1,2$) versus the normalized exact DO coefficients $Y_i^{DO}$ at four different times $t=0,\pi/3, 2\pi/3$ and $\pi$. The random coefficients as functions of the random variable $\xi$ evolve with time and develop a subtle wavy structure, while preserving the orthogonality. The NN-DO method uncovers the evolution behavior of $Y_i$. Figure~\ref{fig:adv_do_mean_var} shows the mean and variance of the NN-DO solution versus the exact ones, at $t=2\pi/3$ and $t=\pi$. Apparently, the scale of variance is large compared to the scale of mean, indicating that the random fluctuation dominates the averaged solution profile. Table~\ref{tab:adv_do} summarizes the $L_2$ error (defined by $\|f_{NN}-f_{exact}\|_{2}$ for any function $f$) and the relative $L_2$ error (defined by $\|f_{NN}-f_{exact}\|_{2}/\|f_{exact}\|_{2}$) of the NN-DO results versus the exact solutions at the final time $T=\pi$, indicating the good performance of the NN-DO method.

\begin{table}[htbp]
\centering
\begin{tabular}{|c|c|c|c|c|c|c|c|c|}
\hline
    & $\bm{\E[u]}$ & \textbf{Var}[$\bm{u}$] & $\bm{a_1}$ & $\bm{a_2}$ & $\bm{u_1}$ & $\bm{u_2}$ & $\bm{Y_1}$ & $\bm{Y_2}$\\
\hhline{|=|=|=|=|=|=|=|=|=|}
$L_2$ error & 0.0006 & 0.0006 & 0.0010 & 0.0051 & 0.0001 & 0.0002 & 0.0007 & 0.0013 \\\hline
Relative $L_2$ error & 1.96$\%$ & 0.11$\%$ & 0.09$\%$ & 0.55$\%$ & 0.04$\%$ & 0.04$\%$ & 0.52$\%$ & 0.93$\%$ \\\hline
\end{tabular}
\caption{Stochastic advection equation (NN-DO). The $L_2$ and relative $L_2$ errors of NN-DO solutions versus the exact solutions at the final time $T=\pi$.}\label{tab:adv_do}
\end{table}

\subsubsection{Case 2: NN-BO Method}
We solve the same problem (Eq.~\ref{eqn:advection}) again, but this time we use the BO constraints by including Eq.~\ref{eqn:bo_loss} as part of the loss function. The initial conditions and reference solutions for the BO components, i.e., $a_i$, $u_i$ and $Y_i$, are the same as those of the previous case, and the neural networks used to approximate the BO components have the same size with the networks used in the previous case.

\begin{figure}[htbp]
    \centering
    \begin{subfigure}{.5\textwidth}
        \centering
        \includegraphics[width=\linewidth]{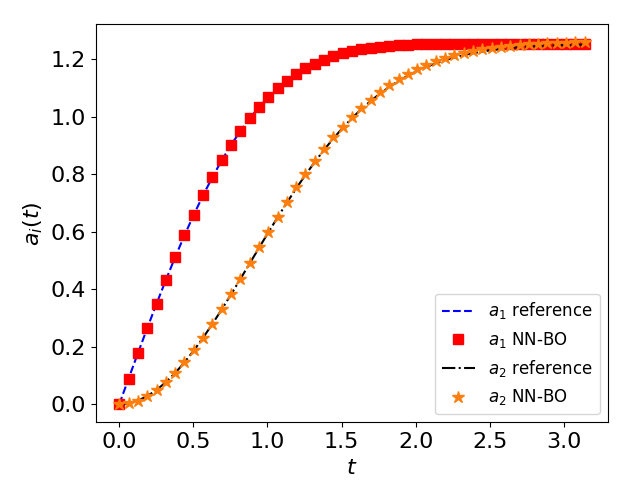}
        \caption{}\label{fig:adv_bo_ai}
    \end{subfigure}%
    \begin{subfigure}{.5\textwidth}
        \centering
        \includegraphics[width=\linewidth]{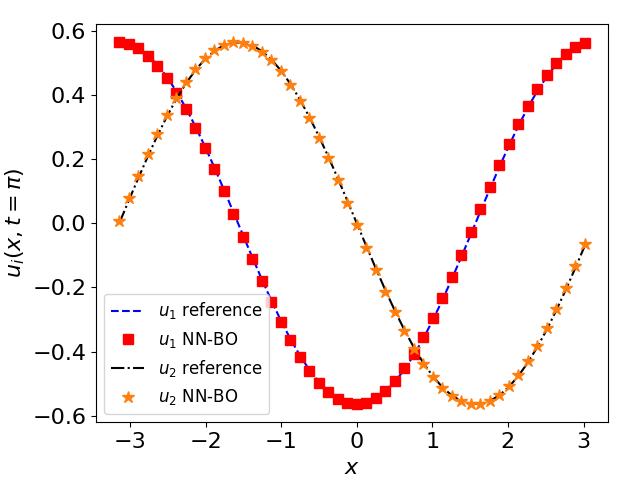}
        \caption{}\label{fig:adv_bo_ui}
    \end{subfigure}
    \caption{Stochastic advection equation (NN-BO). Left: Plot of the evolution of the scaling factors $a_i$; Right: Plot of the bases $u_i$ at the final time $T=\pi$ versus the exact solutions. The scattered points for $u_i$ denote the training points in the physical space.}
\end{figure}

\begin{figure}[htbp]
    \centering
    \begin{subfigure}{.5\textwidth}
        \centering
        \includegraphics[width=\linewidth]{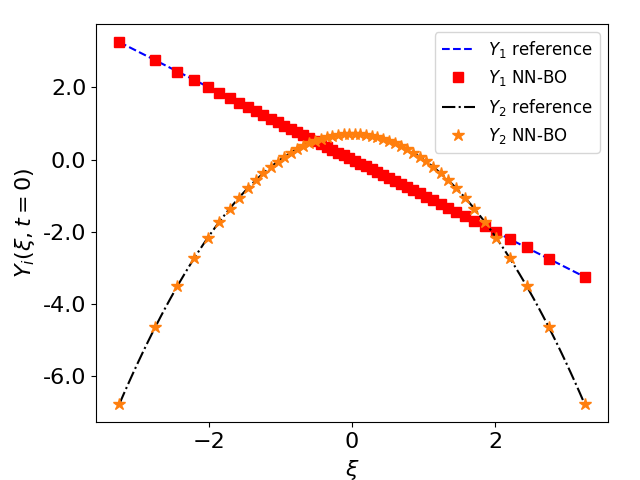}
        \caption{}
    \end{subfigure}%
    \begin{subfigure}{.5\textwidth}
        \centering
        \includegraphics[width=\linewidth]{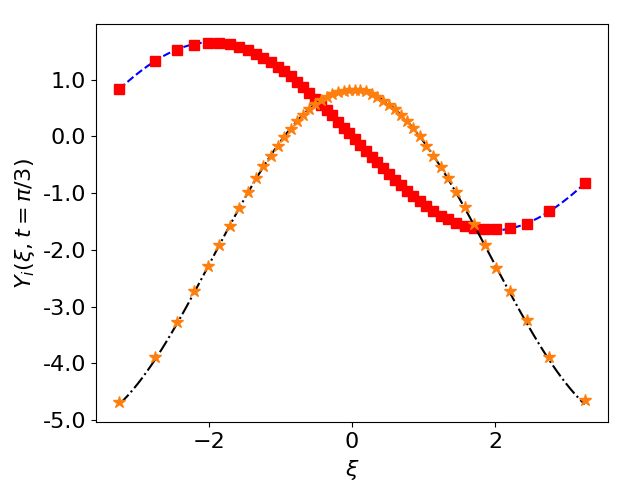}
        \caption{}
    \end{subfigure}
    \begin{subfigure}{.5\textwidth}
        \centering
        \includegraphics[width=\linewidth]{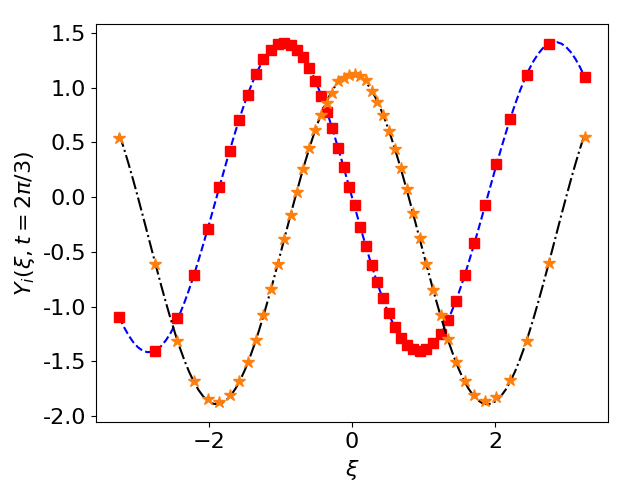}
        \caption{}
    \end{subfigure}%
    \begin{subfigure}{.5\textwidth}
        \centering
        \includegraphics[width=\linewidth]{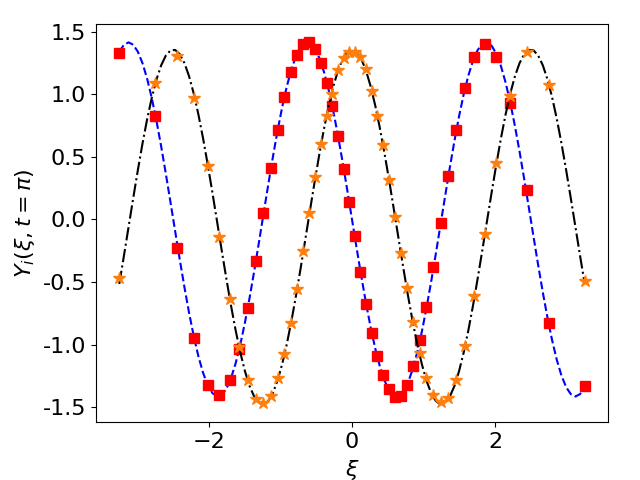}
        \caption{}
    \end{subfigure}
    \caption{Stochastic advection equation (NN-BO). Solutions for the random coefficients $Y_1$ and $Y_2$ at four different times $t=0, \pi/3, 2\pi/3 \text{ and } \pi$. Both of them agree with the exact solutions. The scattered points denote the training points in the probabilistic space.}\label{fig:adv_bo_yi}
\end{figure}

\begin{figure}[htbp]
    \centering
    \begin{subfigure}{.5\textwidth}
        \centering
        \includegraphics[width=\linewidth]{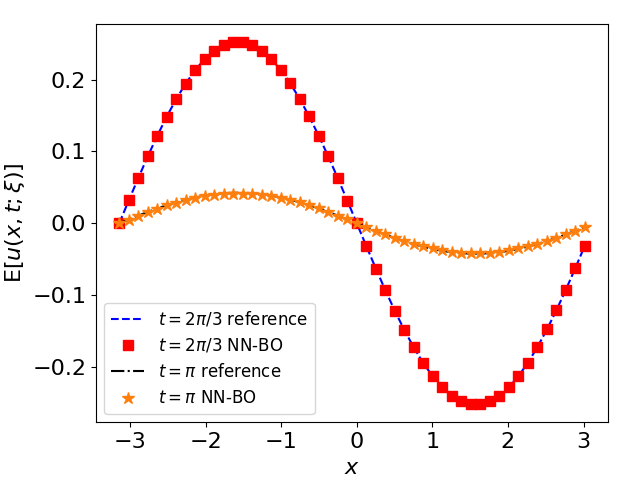}
        \caption{}\label{fig:adv_bo_mean}
    \end{subfigure}%
    \begin{subfigure}{.5\textwidth}
        \centering
        \includegraphics[width=\linewidth]{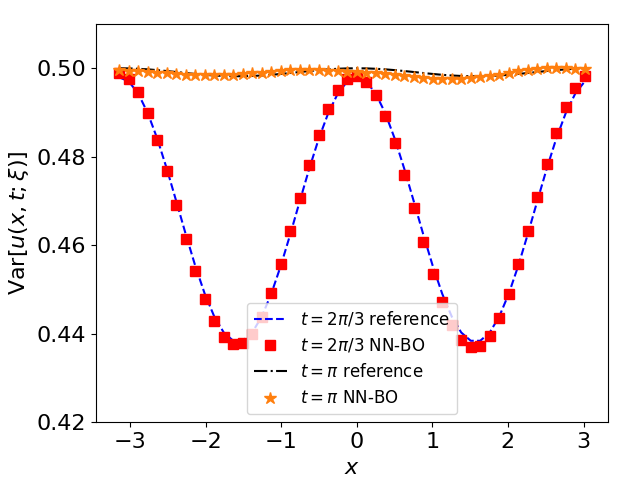}
        \caption{}\label{fig:adv_bo_var}
    \end{subfigure}
    \caption{Stochastic advection equation (NN-BO). Mean and variance of the solution at time $T=2\pi/3$ and $T=\pi$. The scattered points denote the training points in the physical space.}\label{fig:adv_bo_mean_var}
\end{figure}

Similarly, we compare the results obtained using the NN-BO method with the exact solutions. Figure~\ref{fig:adv_bo_ai} and Figure~\ref{fig:adv_bo_ui} display the scaling factors $a_i$ ($i=1,2$) at $t\in[0,\pi]$ and the BO bases $u_i$ ($i=1,2$) at $t=\pi$, respectively. Figure~\ref{fig:adv_bo_yi} shows the stochastic coefficients $Y_i$ ($i=1,2$) versus the normalized exact BO coefficients $Y_i^{BO}$ at four different times: $t=0,\pi/3, 2\pi/3$ and $\pi$. Figure~\ref{fig:adv_bo_mean_var} shows the mean and variance calculated by the NN-BO method at $t = 2\pi/3$ and $t=\pi$. They all show good agreement of the BO solutions with the exact reference solutions. Table~\ref{tab:adv_bo} summarizes the errors of the BO components at the final time $T=\pi$. The NN-BO method demonstrates very good performance similar to the NN-DO method.

\begin{table}[ht]
\centering
\begin{tabular}{|c||c|c|c|c|c|c|c|c|}
\hline
    & $\bm{\E[u]}$ & \textbf{Var}[$\bm{u}$] & $\bm{a_1}$ & $\bm{a_2}$ &  $\bm{u_1}$ & $\bm{u_2}$ & $\bm{Y_1}$ & $\bm{Y_2}$\\
\hhline{|=#=|=|=|=|=|=|=|=|}
$\bm{L_2}$ \textbf{error} & 0.0006 & 0.0006 & 0.0009  & 0.0054 & 0.0051 & 0.0047 & 0.0019 & 0.0019 \\ \hline
\textbf{Relative} $\bm{L_2}$ \textbf{error} & 1.98$\%$ & 0.13$\%$ & 0.08$\%$ & 0.59$\%$ & 1.27$\%$ & 1.18$\%$ & 1.33$\%$ & 1.36$\%$\\ \hline
\end{tabular}
\caption{Stochastic advection equation (NN-BO). The $L_2$ and relative $L_2$ errors of NN-BO solutions versus the exact solutions at the final time $T=\pi$.}\label{tab:adv_bo}
\end{table}

\begin{figure}[htbp]
  \centering
  \includegraphics[width=0.6\linewidth]{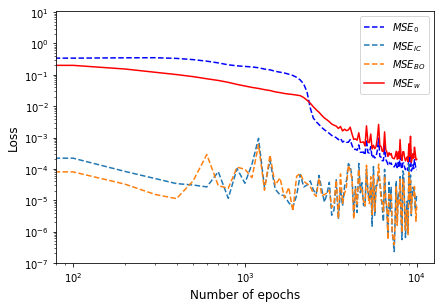}
  \caption{Stochastic advection equation (NN-BO). Various components of the loss function during the training process.}\label{fig:loss_variation}
\end{figure}

To illustrate the effectiveness of the choice of weights in Eq.~\ref{eqn:loss}, we plot the value for each component of the loss function during the first 100000 epochs of training in Figure~\ref{fig:loss_variation}. The decay of the loss associated with the weak formulation $\text{MSE}_\text{w}$ (from more than 0.1 to less than 0.001) is the main effect of the training process. The loss for the initial conditions ($\text{MSE}_\text{IC}$) and the BO conditions ($\text{MSE}_\text{BO}$) are kept small (around $10^{-5}$ to $10^{-4}$), which shows that the whole training process is governed by the initial condition and the BO condition. The loss associated with the original equation ($\text{MSE}_0$) is decaying, indicating that the result is getting closer to the desired solution, but due to the small weight, the contribution of this loss to the total loss is very limited.

\subsection{Application to Nonlinear Stochastic Problem}
In this section, we apply the NN-DO/BO methods to solve nonlinear stochastic problems by considering the following stochastic Burgers' equation:
\begin{equation}\label{eqn:burgers}
\pdv{u}{t} + u \pdv{u}{x} = \nu \pdv[2]{u}{x} + f(x,t;\omega), \quad \forall t \in [0, T] \quad \text{and} \quad x \in \D,
\end{equation}
where the physical domain $\D$ is $[-\pi,\pi]$, and $\nu=0.1$ is the viscosity coefficient. Suppose that the random forcing term $f(x,t;\omega)$ is parameterized by two identically independent uniformly distributed random variables in $[0,1]$, denoted by $\xi_1(\omega)$ and $\xi_2(\omega)$. Then, the stochastic behavior of solution $u(x,t;\omega)$ can be fully described by $\xi_1$ and $\xi_2$, too. In this example, we create a manufactured solution $u(x,t;\xi_1,\xi_2)$ such that the exact DO and BO components can be calculated explicitly. The manufactured solution is
\begin{equation}\label{eqn:burgers_sln}
\begin{aligned}
    u(x,t;\xi_1,\xi_2) = &-\sin(x-t) - \sqrt{3}(1.5+\sin(t))\cos(x-t)(2\xi_1-1)\\
    &+ \sqrt{3}(1.5+\cos(3t))\cos(2x-3t)(2\xi_2-1).
\end{aligned}
\end{equation}
The random forcing term $f(x,t;\omega)$ can be calculated given the manufactured solution. Due to its lengthy expression, here we omit writing down the explicit formula for $f(x,t;\omega)$. Without going into too much detail, Eq.~\ref{eqn:burgers_sln} can be rewritten as either a DO expansion or a BO expansion, given by:
\begin{itemize}
    \item DO components:
    \begin{equation}
        u(x,t;\xi_1, \xi_2) = \E[u](x,t) + u_1^{DO}(x,t)Y_1^{DO}(t; \xi_1, \xi_2) + u_2^{DO}(x,t)Y_2^{DO}(t; \xi_1, \xi_2),
    \end{equation}
    where
    \begin{equation}\label{eqn:bur_do_exact}
        \begin{gathered}
        u_1^{DO}(x,t) = -\frac{1}{\sqrt{\pi}}\cos(x-t),\, u_2^{DO}(x,t) = \frac{1}{\sqrt{\pi}}\cos(2x-3t),\\
        Y_1^{DO}(t;\xi_1,\xi_2) = \sqrt{3\pi}(1.5+\sin(t))(2\xi_1-1),\\
        Y_2^{DO}(t;\xi_1,\xi_2) = \sqrt{3\pi}(1.5+\cos(3t))(2\xi_2-1);
        \end{gathered}
    \end{equation}
    \item BO components:
    \begin{equation}
        u(x,t;\xi_1, \xi_2) = \E[u](x,t) + u_1^{BO}(x,t)Y_1^{BO}(t; \xi_1, \xi_2) + u_2^{BO}(x,t)Y_2^{BO}(t; \xi_1, \xi_2),
    \end{equation}
    where
    \begin{equation}\label{eqn:bur_bo_exact}
        \begin{gathered}
        u_1^{BO}(x,t) = -(1.5+\sin(t))\cos(x-t),\\
        u_2^{BO}(x,t) = (1.5+\cos(3t))\cos(2x-3t),\\
        Y_1^{BO}(t;\xi_1,\xi_2) = \sqrt{3}(2\xi_1-1),\, Y_2^{BO}(t;\xi_1,\xi_2) = \sqrt{3}(2\xi_2-1).
        \end{gathered}
    \end{equation}
\end{itemize}

If we normalize the bases and the random coefficients, and write the above expansions in the form of Eq.~\ref{eqn:expan}, both the DO expansion and the BO expansion yield the same expression:
\begin{equation}\label{eqn:bur_expansion}
    \begin{aligned}
    &u_1(x,t) = -\frac{1}{\sqrt{\pi}}\cos(x-t),\, &&u_2(x,t) = \frac{1}{\sqrt{\pi}}\cos(2x-3t),\\
    &a_1(t) = \sqrt{\pi}(1.5+\sin(t)), &&a_2(t) = \sqrt{\pi}(1.5+\cos(3t)),\\
    &Y_1(t;\xi_1,\xi_2) = 2\xi_1-1, &&Y_2(t;\xi_1,\xi_2) = 2\xi_2-1.
    \end{aligned}
\end{equation}

We obtain the solution until $T=10\pi$ to demonstrate the long-term performance of the NN-DO/BO method. In practice, we divide the time domain into ten non-overlapping subdomains of equal length, each of which has the length $\pi$. In each subdomain the components of Eq.~\ref{eqn:expan} are approximated by an independent set of feed-forward neural networks. We train the time domains one-after-another and use the results from the previous interval at the end time as the initial conditions for the next subdomain. This domain decomposition strategy circumvents the difficulty of approximating functions of massive fluctuations with a single neural network, and thus will make the training process easier. We use an equal number of training points for all time subdomains, and set $n_t = 30$ and $n_x = 50$. Again, the samples of $\{t_c^s\}_{s=1}^{n_t}$ are drawn from a uniform distribution, and the spatial training points $\{x_c^k\}_{k=1}^{n_x}$ are equidistantly distributed in $[-\pi, \pi]$. For the training points in the stochastic space, we use eighth-order Gauss-Legendre quadrature rule for both $\xi_1$ and $\xi_2$, generating 64 points in the probabilistic space. The same neural network setups are implemented for the following two test cases: the $\overbar{u}_{nn}$, $A_{nn}$ and $Y_{nn}$ networks all have three hidden layers, each of which has 32 neurons, and the $U_{nn}$ network is constructed with 3 hidden layers and 64 neurons per hidden layer. We only change the loss function in favor of either the DO or the BO condition. The neural networks are trained with an Adam optimizer (learning rate 0.001) for 50000 epochs.

\subsubsection{Case 1: NN-DO Method}

\begin{figure}[htbp]
    \centering
    \includegraphics[width=0.9\linewidth]{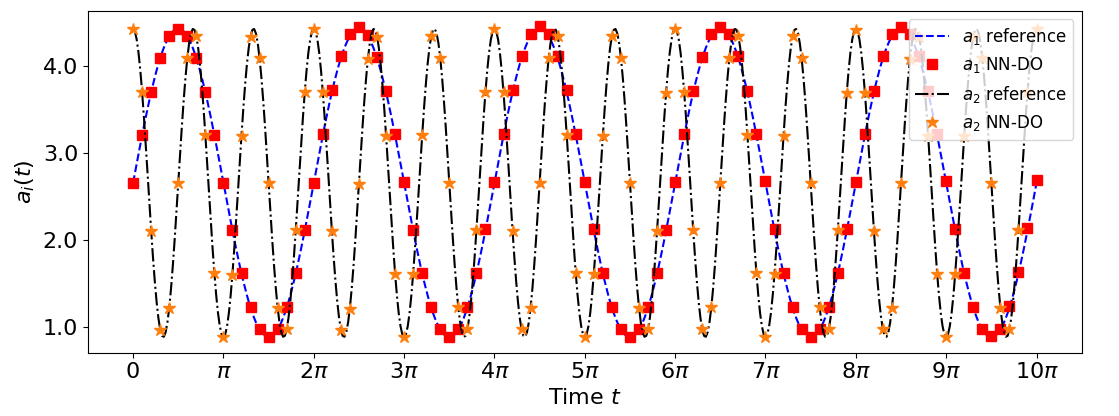}
    \caption{Stochastic Burgers' equation (NN-DO). A comparison of neural network approximations and exact solutions of the scaling factors $a_i$ ($i=1,2$), as functions of $t$ ($t\in[0,10\pi]$).}\label{fig:bur_do_ai}
\end{figure}

\begin{figure}[htbp]
    \centering
    \begin{subfigure}{.5\textwidth}
        \centering
        \includegraphics[width=\linewidth]{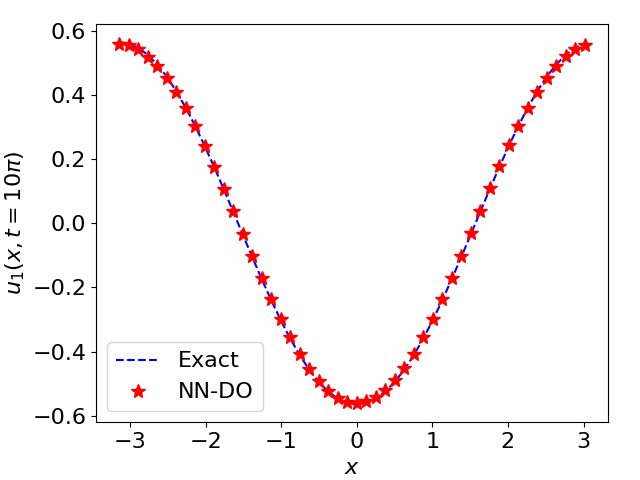}
        \caption{}
    \end{subfigure}%
    \begin{subfigure}{.5\textwidth}
        \centering
        \includegraphics[width=\linewidth]{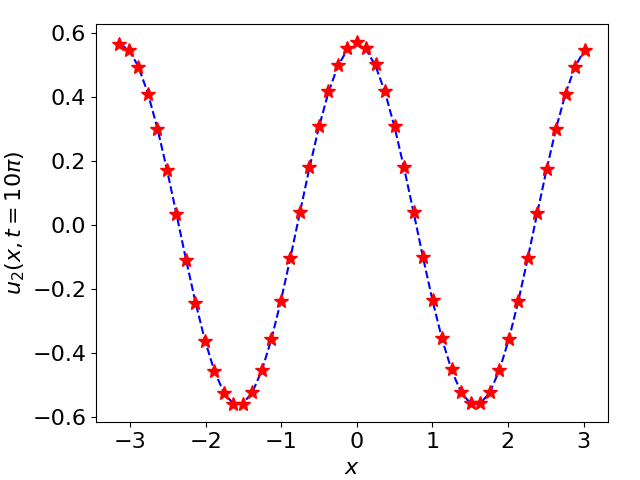}
        \caption{}
    \end{subfigure}
    \caption{Stochastic Burgers' equation (NN-DO). A comparison of neural network approximations and exact solutions of the bases $u_i$ ($i=1,2$) at the final time $t=10\pi$. The red stars denote the training points in the physical space.}\label{fig:bur_do_ui}
\end{figure}

\begin{figure}[htbp]
    \centering
    \begin{subfigure}{.5\textwidth}
        \centering
        \includegraphics[width=\linewidth]{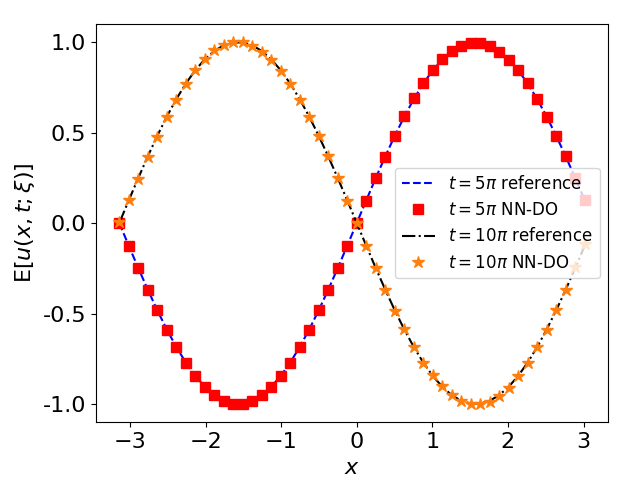}
        \caption{}\label{fig:bur_do_mean}
    \end{subfigure}%
    \begin{subfigure}{.5\textwidth}
        \centering
        \includegraphics[width=\linewidth]{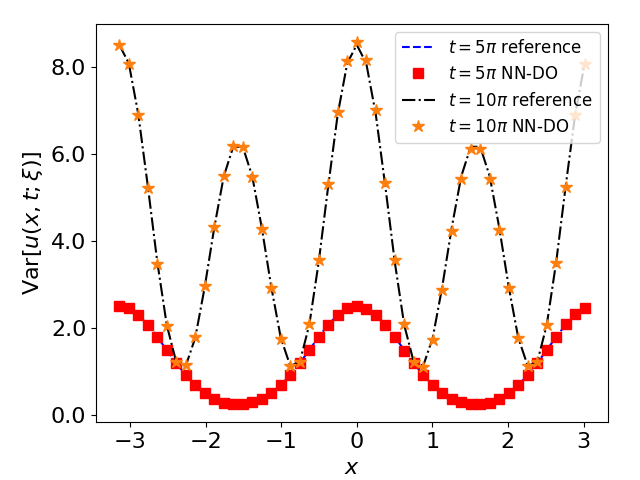}
        \caption{}\label{fig:bur_do_var}
    \end{subfigure}
    \caption{Stochastic Burgers' equation (NN-DO). Mean and variance of the solution at time $t=5\pi$ and $t=10\pi$, obtained using the NN-DO method. Both of them show good agreement with the reference exact value. The scattered points denote the training points in the physical space.}\label{fig:bur_do_mean_var}
\end{figure}

\begin{figure}[htbp]
    \centering
    \includegraphics[width=0.8\linewidth]{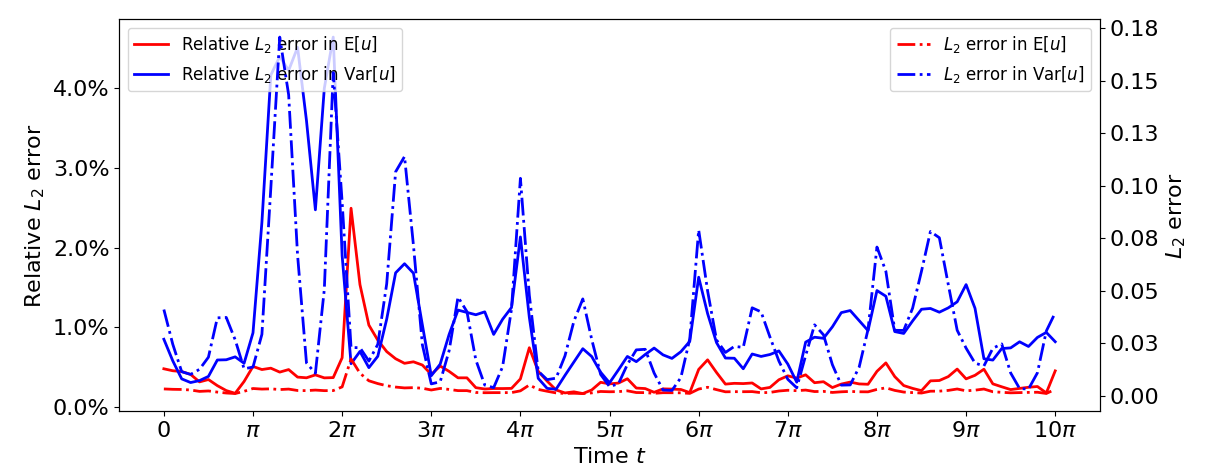}
    \caption{Stochastic Burgers' equation (NN-DO). The $L_2$ and relative $L_2$ errors in the mean and variance obtained by the NN-DO method. The relative error in variance is slightly higher than the relative error in mean, and both errors are below $2\%$ for most of the time.}\label{fig:bur_do_error}
\end{figure}

First, we test the NN-DO method. The initial conditions are taken directly from Eq.~\ref{eqn:bur_expansion}. Figure~\ref{fig:bur_do_ai} shows the evolution of $a_i$ ($i=1,2$) as time grows, where the low frequency component, $a_1$, and the high frequency component, $a_2$, co-exist at the same amplitude. They do not decay with time, indicating that the stochasticity in the system has already reached a fully developed state. In Figure~\ref{fig:bur_do_ui}, we compare the bases $u_i$ ($i=1,2$) at $t=10\pi$ obtained from the NN-DO method to the exact solutions, and in Figure~\ref{fig:bur_do_mean_var}, we plot the NN-DO solution mean and variance at two times, $t=10\pi$ and $t=5\pi$, versus the exact values. It is evident that the NN-DO solutions agree with the exact reference solutions very well. From Figure~\ref{fig:bur_do_var} we can observe that the solution variance evolves from $t=5\pi$ to $t=10\pi$ to develop a greater magnitude range and a more complex shape, and the NN-DO method precisely captures this progress. Figure~\ref{fig:bur_do_error} shows the relative $L_2$ errors of the solution mean and variance, and in Table~\ref{tab:bur_do} we report both errors for all the DO components at the final time $T=10\pi$. All relative $L_2$ errors are around or less than $1\%$, indicating the good performance of the proposed NN-DO method.

\begin{table}[ht]
\centering
\begin{tabular}{|c||c|c|c|c|c|c|c|c|}
\hline
    & $\bm{\E[u]}$ & \textbf{Var}[$\bm{u}$] & $\bm{a_1}$ & $\bm{a_2}$ &  $\bm{u_1}$ & $\bm{u_2}$ & $\bm{Y_1}$ & $\bm{Y_2}$ \\
\hhline{|=#=|=|=|=|=|=|=|=|}
$\bm{L_2}$ \textbf{error} & 0.0029 & 0.0278  & 0.0104 & 0.0084 & 0.0042 & 0.0021 & 0.0008 & 0.0004 \\ \hline
\textbf{Relative} $\bm{L_2}$ \textbf{error} & 0.40$\%$ & 0.57$\%$ & 0.35$\%$ & 0.28$\%$ & 1.04$\%$ & 0.53$\%$ & 0.62$\%$ & 0.34$\%$ \\ \hline
\end{tabular}
\caption{Stochastic Burgers' equation (NN-DO). The $L_2$ and relative $L_2$ errors of NN-DO solutions versus the exact solutions at the final time $T=10\pi$.}\label{tab:bur_do}
\end{table}

\subsubsection{Case 2: NN-BO Method}
\begin{figure}[htbp]
    \centering
    \includegraphics[width=0.9\linewidth]{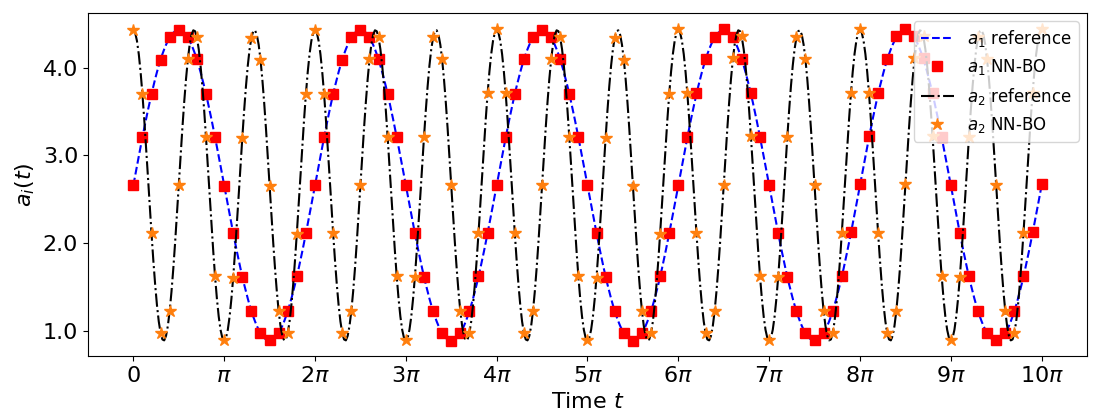}
    \caption{Stochastic Burgers' equation (NN-BO). A comparison of neural network approximations and exact solutions of the scaling factors $a_i$ ($i=1,2$) as functions of $t$ ($t\in[0,10\pi]$). They correspond to the eigenvalues in the standard BO method. As we can see, there is a significant amount of eigenvalue crossings during the whole time evolution and also within each subdomain. Therefore, the standard BO method cannot be directly applied to this problem.}\label{fig:bur_bo_ai}
\end{figure}

\begin{figure}[htbp]
    \centering
    \begin{subfigure}{.5\textwidth}
        \centering
        \includegraphics[width=\linewidth]{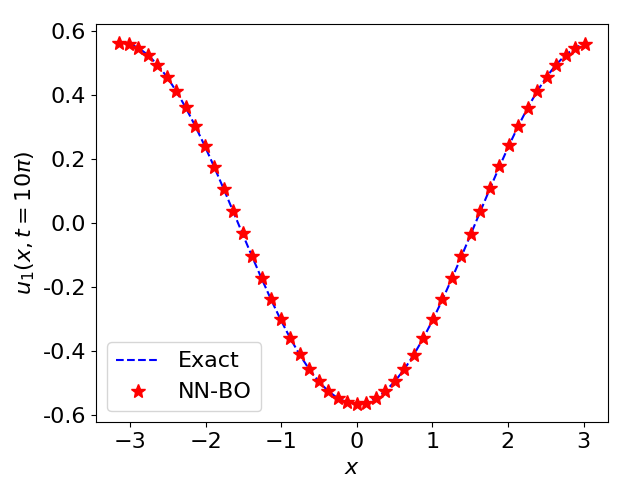}
        \caption{}
    \end{subfigure}%
    \begin{subfigure}{.5\textwidth}
        \centering
        \includegraphics[width=\linewidth]{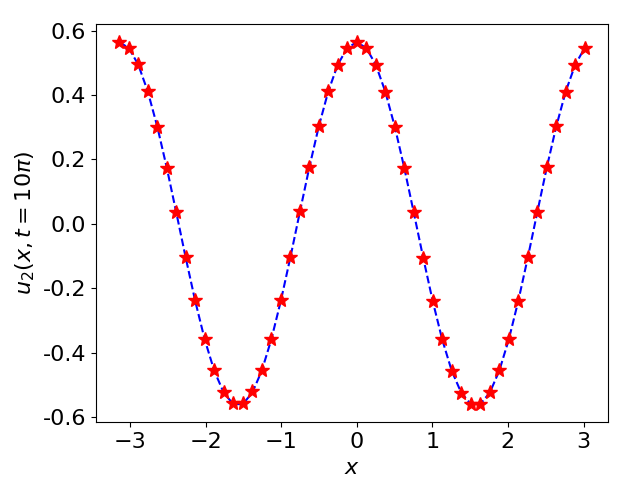}
        \caption{}
    \end{subfigure}
    \caption{Stochastic Burgers' equation (NN-BO). A comparison of neural network approximations and exact solutions of the bases $u_i$ ($i=1,2$) at the final time $t=10\pi$. The red stars denote the training points in the physical space.}\label{fig:bur_bo_ui}
\end{figure}

\begin{figure}[htbp]
    \centering
    \begin{subfigure}{.5\textwidth}
        \centering
        \includegraphics[width=\linewidth]{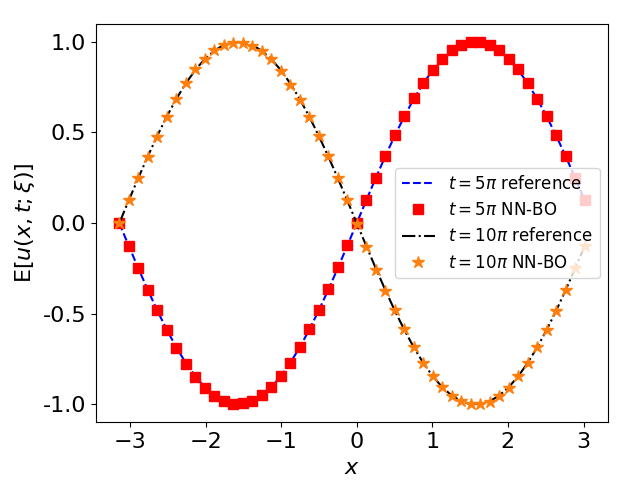}
        \caption{}\label{fig:bur_bo_mean}
    \end{subfigure}%
    \begin{subfigure}{.5\textwidth}
        \centering
        \includegraphics[width=\linewidth]{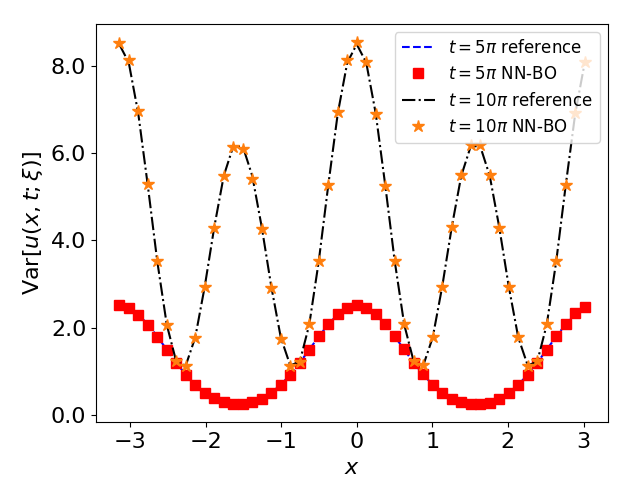}
        \caption{}\label{fig:bur_bo_var}
    \end{subfigure}
    \caption{Stochastic Burgers' equation (NN-BO). Mean and variance of the solutions at time $t=5\pi$ and $t=10\pi$, obtained using the NN-BO method. Both of them show good agreement with the reference exact values.}\label{fig:bur_bo_mean_var}
\end{figure}

\begin{figure}[htbp]
    \centering
    \includegraphics[width=0.8\linewidth]{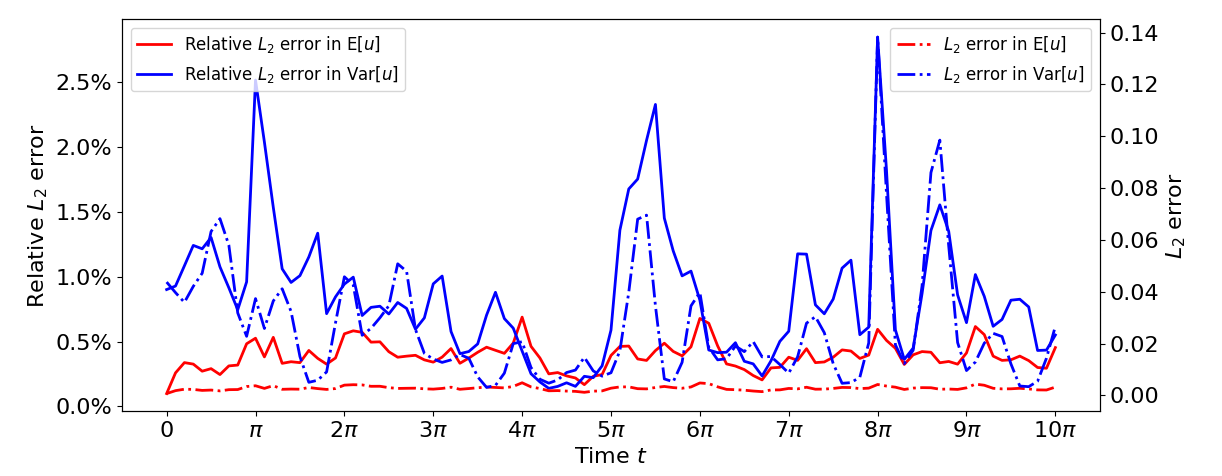}
    \caption{Stochastic Burgers' equation (NN-BO). The $L_2$ and relative $L_2$ errors in the mean and variance calculated by the NN-BO method. The relative error in variance is slightly higher than the relative error in mean, and both errors are below $2\%$ for most of the time.}\label{fig:bur_bo_error}
\end{figure}

In this section we use the BO constraints to train the neural networks. Similar to its NN-DO counterpart, here we provide all the figures (Figure~\ref{fig:bur_bo_ai}--Figure~\ref{fig:bur_bo_error}) showing a comparison between the NN-BO results and the reference exact solutions. To avoid redundancy, we refer the readers to read the captions below the figures and will skip explaining each of them one-by-one. However, we would like to note that in Figure~\ref{fig:bur_bo_ai}, the scaling factors $a_i$ correspond to the eigenvalues in the standard BO method, and there is a significant amount of eigenvalue crossings during the whole time evolution, and also within each time subdomain. In this situation, the standard BO method would fail due to the lack of explicit formulas for matrices $M$ and $S$ in Eq.~\ref{eqn:S_M}. The proposed NN-BO method does not suffer from this issue. In Table~\ref{tab:bur_bo} we report both the $L_2$ and relative $L_2$ errors for all the BO components at the final time $T=10\pi$. As with the NN-DO method, all relative $L_2$ errors are less than $1\%$, indicating the good performance of the NN-BO method.

\begin{table}[ht]
\centering
\begin{tabular}{|c||c|c|c|c|c|c|c|c|}
\hline
    & $\bm{\E[u]}$ & \textbf{Var}[$\bm{u}$] & $\bm{a_1}$ & $\bm{a_2}$ &  $\bm{u_1}$ & $\bm{u_2}$ & $\bm{Y_1}$ & $\bm{Y_2}$ \\
\hhline{|=#=|=|=|=|=|=|=|=|}
$\bm{L_2}$ \textbf{error} & 0.0032 & 0.0267 & 0.0055 & 0.0073 &  0.0018  &   0.0020 &  0.0007 & 0.0005 \\ \hline
\textbf{Relative} $\bm{L_2}$ \textbf{error} & 0.45$\%$ & 0.55$\%$  &  0.19$\%$  &  0.25$\%$ &  0.45$\%$ & 0.49$\%$ &  0.59$\%$ & 0.39$\%$ \\ \hline
\end{tabular}
\caption{Stochastic Burgers' equation (NN-BO). The $L_2$ and relative $L_2$ errors of NN-BO solutions versus the exact solutions at the final time $T=10\pi$.}\label{tab:bur_bo}
\end{table}

\subsection{Application to Nonlinear Diffusion-Reaction Equation}
Consider the following reaction diffusion equation with a nonlinear source term:
\begin{equation}\label{eqn:reac_diff}
\pdv{u}{t} = a u_{xx} + bu^2 + f(x;\omega), \quad \forall t \in [0, 1] \text{ and } x \in [-1, 1],
\end{equation}
where the random force $f(x;\omega)=(1-x^2)g(x;\omega)$ is the source of randomness, while $a$ and $b$ are time-independent diffusion and reaction coefficients, respectively. The random process $g(x;\omega)$ is modeled as a Gaussian random field, i.e., $g(x;\omega) \sim \mathcal{GP}(1, C(x_1, x_2))$, where $C(x_1, x_2)$ is a squared exponential kernel with standard deviation $\sigma_g$ and correlation length $l_c$:
\begin{equation}
    C(x_1, x_2) = \sigma_g^2\exp \left(-\frac{(x_1-x_2)^2}{l_c^2}\right).
\end{equation}
The solution satisfies the Dirichlet boundary conditions, $u(-1,t;\omega)=u(1,t;\omega)=0$, and the deterministic initial condition $u(x,0;\omega)=-\sin(\pi x)$. We consider two different scenarios here:
\begin{itemize}
    \item Forward problem: the coefficients $a$ and $b$ are given, and we solve for $u(x,t;\omega)$.
    \item Inverse problem: the coefficients $a$ and $b$ are unknown but additional information for $u(x,t;\omega)$ is given; we solve for $u(x,t;\omega)$ while we also aim to identify $a$ and $b$.
\end{itemize}
For brevity, here we only show the results obtained from the NN-BO method as the NN-DO method exhibits a similar performance.

\subsubsection{Forward Problem}
We set the diffusion coefficient $a=0.1$ and the reaction coefficient $b=0.5$. For the random force $f(x;\omega)$, we set $\sigma_g=1$ and $l_c=0.1$, thus requiring 19 KL modes to capture at least $98\%$ of the fluctuation energy of $f(x;\omega)$. The neural networks used in the NN-BO method are built as follows: $\overbar{u}_{nn}$ has three hidden layers with 32 neurons per layer, $U_{nn}$ and $Y_{nn}$ have three hidden layers with 64 neurons per layer, and $A_{nn}$ is composed of $N$ independent neural networks ($N$ is the number of BO expansion terms), each of which has three hidden layers and four neurons per layer, approximating one single scaling factor $a_i$. This is because we expect that $a_i$ may oscillate greatly in vastly different scales during the time evolution. We use $n_x=51$ equidistantly distributed training points $\{x_c^k\}_{k=1}^{n_x}$ in space, $n_t=50$ uniformly distributed training points $\{t_c^s\}_{k=1}^{n_t}$ in the time domain, and $n_l=1000$ random samples $\{\xi_c^l\}_{l=1}^{n_\xi}$ in the 19-dimensional random space. The neural networks are trained with an Adam optimizer (learning rate 0.001) for 300000 epochs.

First, we investigate the performance of NN-BO method using six BO expansion terms. To obtain the reference solution for the BO decomposition, we numerically solved the original BO equations with the finite difference scheme in space and a 3rd-level Adam-Bashforth scheme in time. Due to the deterministic initial condition, in practice we start with a Monte Carlo method until $t=0.01$, and then switch to solving the BO equations. To obtain the reference for the solution statistics we solve the SPDE using a Monte Carlo method with 1000 samples. 

\begin{figure}[htbp]
    \centering
    \begin{subfigure}{.5\textwidth}
        \centering
        \includegraphics[width=\linewidth]{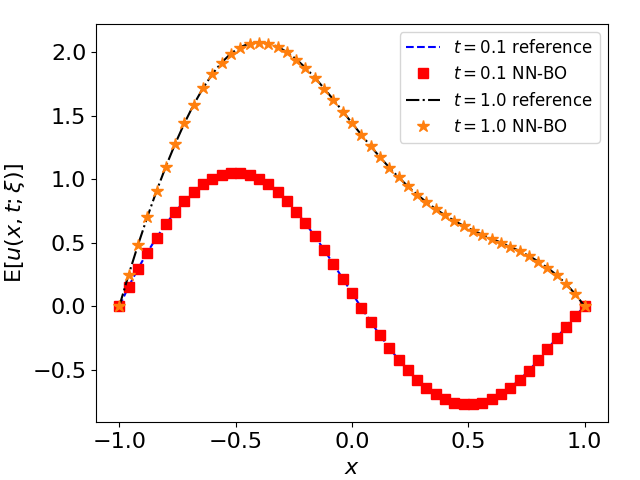}
        \caption{}\label{fig:nonlinear_forward_mean}
    \end{subfigure}%
    \begin{subfigure}{.5\textwidth}
        \centering
        \includegraphics[width=\linewidth]{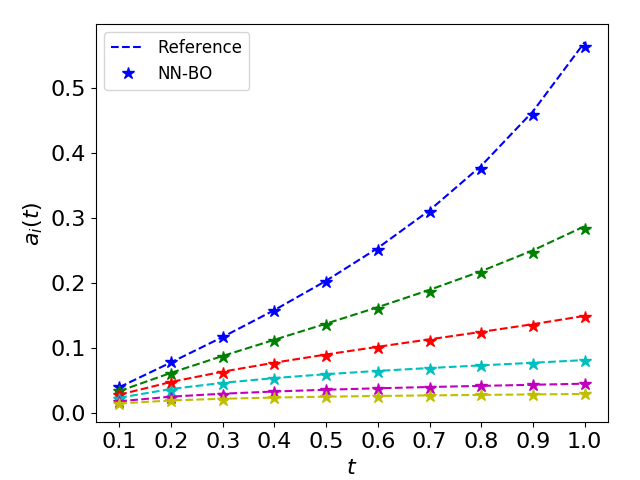}
        \caption{}\label{fig:nonlinear_forward_ai}
    \end{subfigure}
    \caption{Stochastic diffusion-reaction equation (forward problem). Left: Solution mean at $t=0.1$ and $t=1.0$. The reference mean is calculated from a Monte Carlo simulation; Right: Scaling factors $a_i$ at different time steps. The reference $a_i$ are calculated using the standard numerical BO method.}
\end{figure}

\begin{figure}[htbp]
    \centering
    \includegraphics[width=\linewidth]{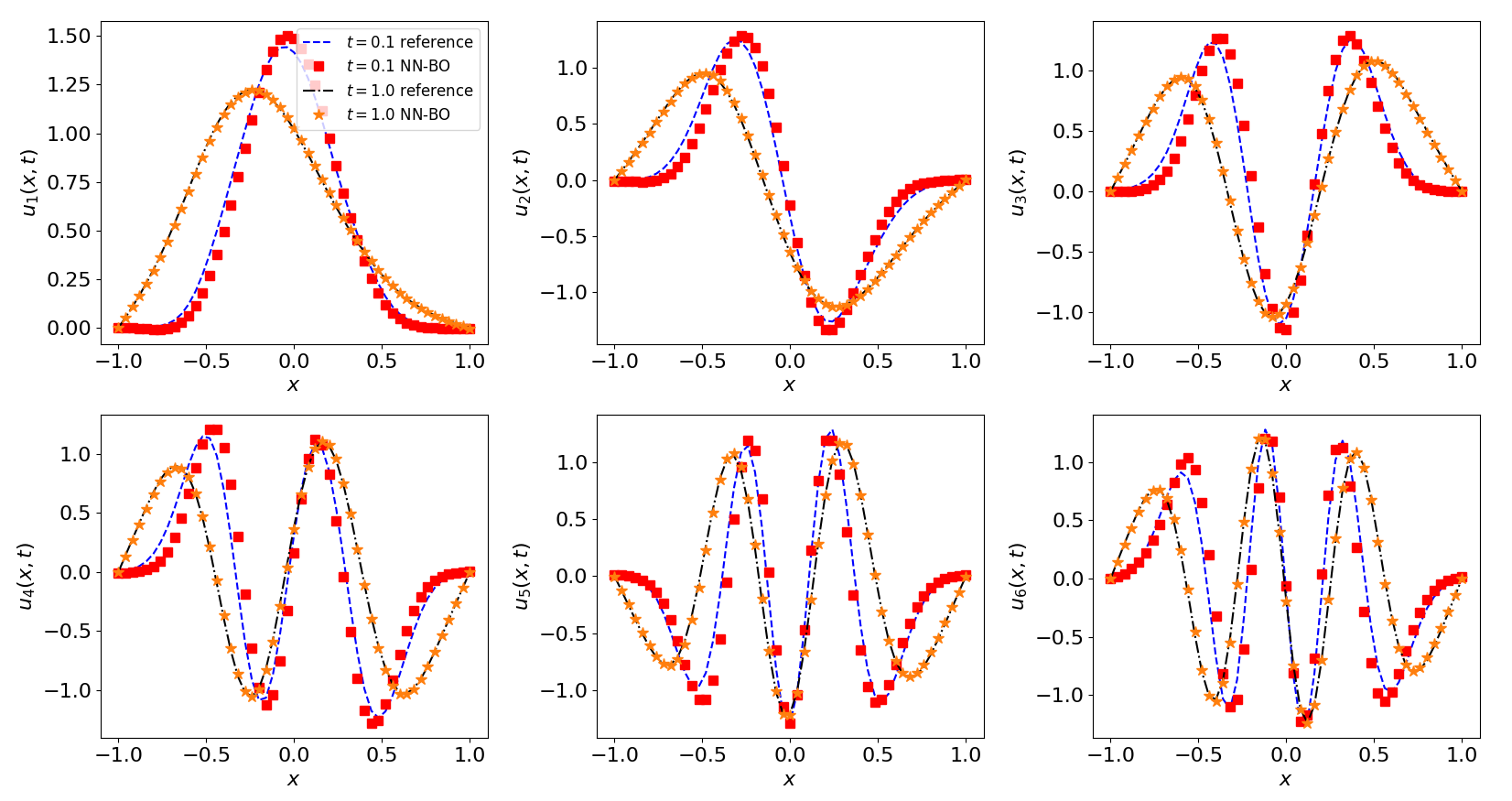}
    \caption{Stochastic diffusion-reaction equation (forward problem). The BO modes $u_i$ at $t=0.1$ and $t=1.0$; the reference $u_i$ are calculated using the standard numerical BO method.}\label{fig:nonlinear_forward_ui}
\end{figure}

Figure~\ref{fig:nonlinear_forward_mean} shows the NN-BO solution mean at $t=0.1$ and $t=1.0$, and Figure~\ref{fig:nonlinear_forward_ai} shows the evolution of the scaling factors $a_i$, where the first four BO modes gradually pick up energy as the result of the nonlinear source term, while the energy in the fifth and sixth modes is relatively stable in time. This illustrates the efficiency of the BO representation, i.e., only a small number of modes is necessary to capture most of the stochasticity in this 19-dimensional SPDE. Figure~\ref{fig:nonlinear_forward_ui} compares the modal functions learned from the NN-BO method with the reference, and Table~\ref{tab:nonlinear_yerror} displays the root mean squared error of the random coefficients $Y_i$. The proposed NN-BO method generates accurate predictions at both the early stage of the solution ($t=0.1$) and the end time ($t=1.0$).

\begin{table}[ht]
\centering
\begin{tabular}{|c||c|c|c|c|c|c|}
\hline
\textbf{RMSE}  & \bm{$Y_1$} & \bm{$Y_2$} & \bm{$Y_3$} & \bm{$Y_4$} & \bm{$Y_5$} & \bm{$Y_6$} \\
\hhline{|=#=|=|=|=|=|=|}
$t=0.1$ & 0.098 & 0.175 & 0.225 & 0.292 & 0.237 & 0.275 \\\hline
$t=1.0$ & 0.042 & 0.039 & 0.045 & 0.050 & 0.061 & 0.057 \\\hline
\end{tabular}
\caption{Stochastic diffusion-reaction equation (forward problem). Root mean squared error of the random coefficients $Y_i$ calculated using the NN-BO method at $t=0.1$ and $t=1.0$; the reference $Y_i$ are calculated using the standard numerical BO method.}\label{tab:nonlinear_yerror}
\end{table}

\begin{figure}[htbp]
    \centering
    \begin{subfigure}{.5\textwidth}
        \centering
        \includegraphics[width=\linewidth]{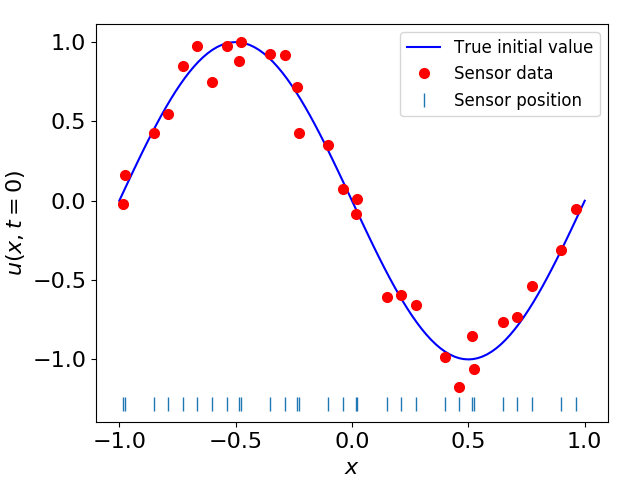}
        \caption{}\label{fig:nonlinear_noisy_ic}
    \end{subfigure}%
    \begin{subfigure}{.5\textwidth}
        \centering
        \includegraphics[width=\linewidth]{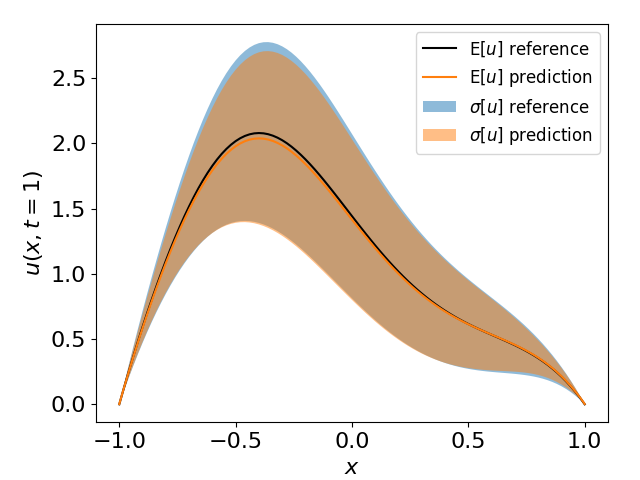}
        \caption{}\label{fig:nonlinear_noisy_solution}
    \end{subfigure}
    \caption{Stochastic diffusion-reaction equation with noisy data as initial condition. Left: Noisy sensor data as initial condition; Right: Mean and standard deviation of the predicted solution $u(x,t;\omega)$ versus the reference mean and standard deviation.}\label{fig:nonlinear_noisy}
\end{figure}

\begin{figure}[htbp]
    \centering
    \begin{subfigure}{.5\textwidth}
        \centering
        \includegraphics[width=\linewidth]{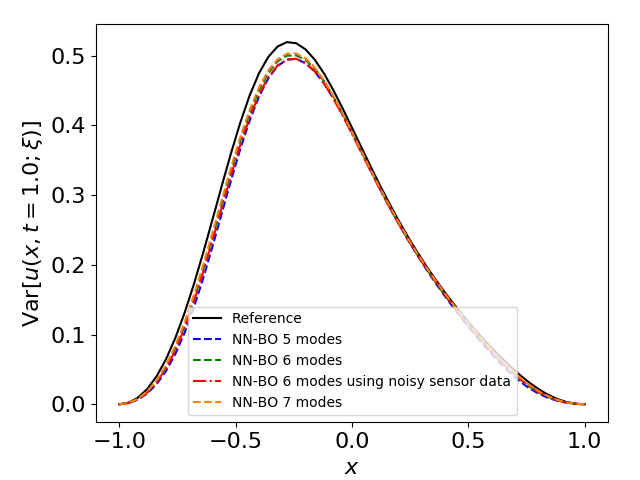}
        \caption{}\label{fig:nonlinear_forward_var}
    \end{subfigure}%
    \begin{subfigure}{.5\textwidth}
        \centering
        \includegraphics[width=\linewidth]{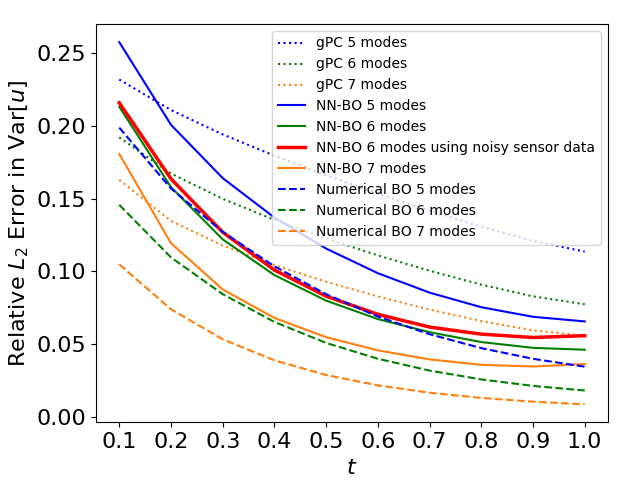}
        \caption{}\label{fig:nonlinear_forward_var_error}
    \end{subfigure}
    \caption{Stochastic diffusion-reaction equation (forward). Left: Variance of the NN-BO solution calculated using 5, 6 and 7 modes; the reference variance is calculated from the Monte Carlo simulation; Right: Comparison of the $L_2$ errors of the solution variance calculated by the NN-BO method, standard numerical BO method and the gPC method. The gPC method generates the largest error since it fails to capture the dynamic evolution of stochastic basis for nonlinear problems.}\label{fig:nonlinear_forward_compare_var}
\end{figure}

Next, we analyze the effect of the number of BO expansion modes by comparing the variances of solution calculated using five, six and seven BO modes, and moreover, we solve the diffusion-reaction equation with noisy sensor data as the initial condition. Figure~\ref{fig:nonlinear_noisy_ic} shows the noisy sensor measurements of $u(x,t=0)$, where the 30 sensors are uniformly placed in the domain, and the red dots are perturbed measurements generated by artificially adding independent Gaussian random noise of standard deviation $0.1$ to the hidden true values. Figure~\ref{fig:nonlinear_noisy_solution} shows a comparison of the NN-BO solution mean and standard deviation, calculated based on noisy sensor data, and the reference mean and standard deviation, obtained with the Monte Carlo simulation. Figure~\ref{fig:nonlinear_forward_var} shows the predicted variance at time $t=1.0$ versus the reference solution. The NN-BO method slightly underestimates the variance due to the truncated expansion, and using noisy sensor measurements as the initial condition does not change the prediction at final time too much. Figure~\ref{fig:nonlinear_forward_var_error} compares the relative $L_2$ error of the solution variance at $t=1.0$ obtained using three different methods: NN-BO, gPC and the standard BO. The gPC method generates the largest error as it fails to capture the evolution of the system's stochastic structure due to the non-linearity, therefore, to achieve the same accuracy, one has to include a larger number of modes using the gPC method than using the BO method. Again, we can observe that better accuracy can be achieved when more modes are included, and we obtain similar accuracy when using noisy sensor data as the initial condition. The NN-BO method is less accurate than the standard numerical BO method due to dominant optimization errors. However, it circumvents the need to generate artificial stochastic initial conditions and can make use of scattered, noisy sensor measurements as constraints, rather than explicit mathematical expressions. Another advantage of the NN-BO method over the standard BO method is that it can solve efficiently a time-dependent nonlinear {\em inverse} stochastic problem.

\subsubsection{Inverse Problem}
Here again we solve Eq.~\ref{eqn:reac_diff} but this time we assume that we do not know the exact diffusion and reaction coefficients $a$ and $b$. Some extra information about $u(x,t;\omega)$ is provided to help us infer these two coefficients. In this example, the extra information is the mean value of $u(x,t;\omega)$ evaluated at three locations $x=-0.5, 0, 0.5$ and at two times $t=0.1, 0.9$, i.e., a total of six measurements of $\E[u]$. We set $\sigma_g=1$ and $l_c=0.4$, and the ``hidden" values of $a$ and $b$ are selected to be $0.5$ and $0.3$, respectively. To solve this inverse problem, we use a BO representation with four modes, and adopt the same setup of the neural networks and training points employed in the forward problem. When setting up the PINNs, $a$ and $b$ are coded as "variables" instead of as "constants" so that they will be tuned at the training stage. Meanwhile, we include an additional term in the loss function that calculates the MSE of the predicted $\overbar{u}_{nn}(x,t)$ versus the measurement data, so that the loss function will make use of the extra information to infer the coefficients. Without loss of generality, we choose both the initial values of $a$ and $b$ to be $1.0$, and in practice these values could be chosen based on reasonable guesses. The neural networks are trained with the Adam optimizer (learning rate 0.001) for 300000 epochs. Same as in the forward problem, the reference solution statistics are calculated with Monte Carlo simulation and the reference BO components are generated by numerically solving the BO equations.

\begin{figure}[htbp]
    \centering
    \begin{subfigure}{.5\textwidth}
        \centering
        \includegraphics[width=\linewidth]{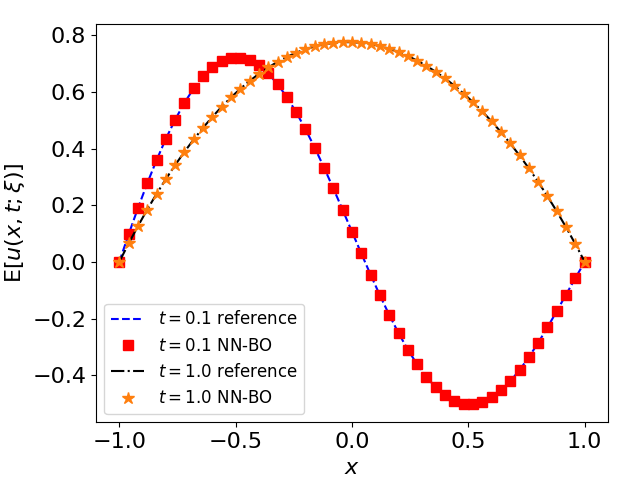}
        \caption{}\label{fig:nonlinear_inverse_mean}
    \end{subfigure}%
    \begin{subfigure}{.5\textwidth}
        \centering
        \includegraphics[width=\linewidth]{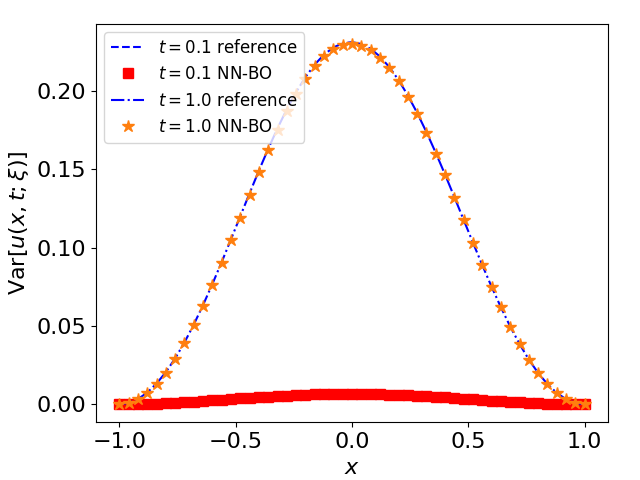}
        \caption{}\label{fig:nonlinear_inverse_var}
    \end{subfigure}
    \caption{Stochastic diffusion-reaction equation (inverse problem). Mean (a) and variance (b) of the NN-BO solution at $t=0.1$ and $t=1.0$; the reference solutions are calculated from the forward problem using the Monte Carlo method.}\label{fig:nonlinear_inverse_mean_var}
\end{figure}

\begin{figure}[htbp]
    \centering
    \includegraphics[width=\linewidth]{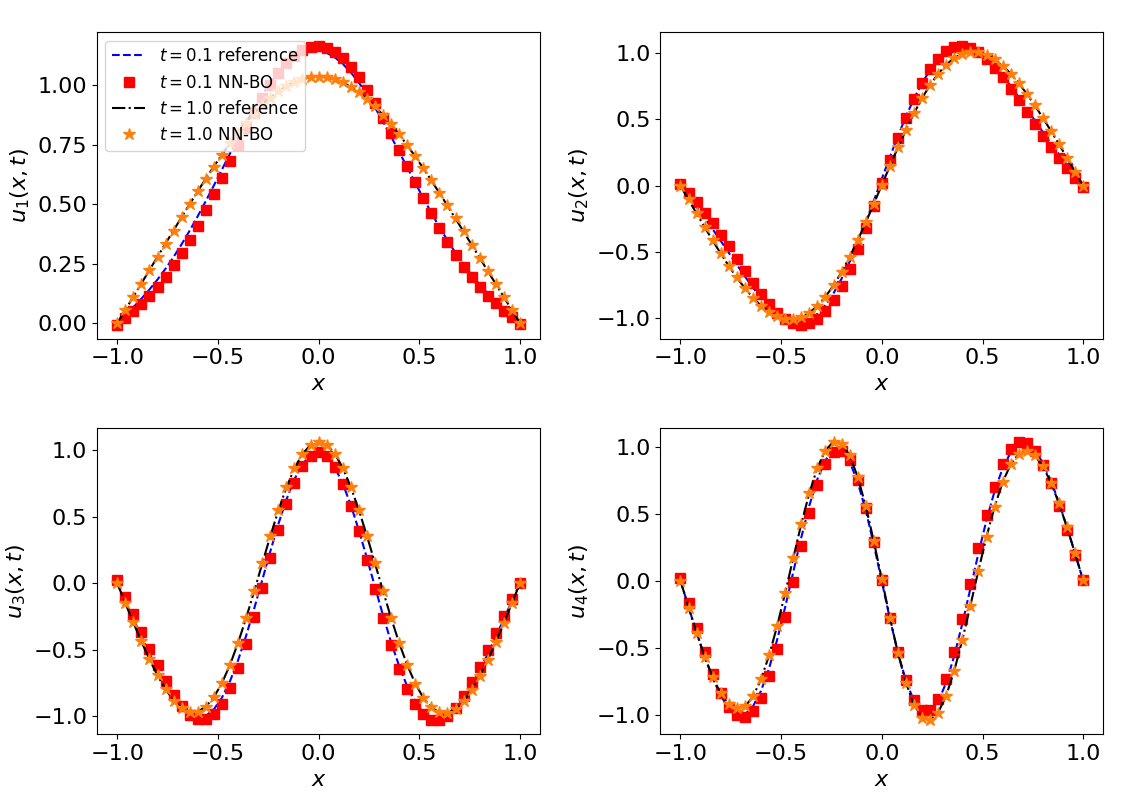}
    \caption{Stochastic diffusion-reaction equation (inverse problem). The BO modes $u_i$ at $t=0.1$ and $t=1.0$; the reference $u_i$ are calculated from the forward problem using the standard numerical BO method.}\label{fig:nonlinear_inverse_ui}
\end{figure}

\begin{table}[htbp]
\centering
\begin{tabular}{|c||c|c|c|c|}
\hline
\textbf{RMSE}  & \bm{$Y_1$} & \bm{$Y_2$} & \bm{$Y_3$} & \bm{$Y_4$} \\\hhline{|=#=|=|=|=|}
$t=0.1$ & 0.061 & 0.088 & 0.075 & 0.084 \\\hline
$t=1.0$ & 0.019 & 0.041 & 0.039 & 0.060 \\\hline
\end{tabular}
\caption{Stochastic diffusion-reaction equation (inverse problem). Root mean squared error of the random coefficients $Y_i$ calculated using the NN-BO method at $t=0.1$ and $t=1.0$; the reference $Y_i$ are calculated from the forward problem using the standard numerical BO method.}\label{tab:inverse_yerror}
\end{table}

\begin{figure}[htbp]
    \centering
    \begin{subfigure}{.5\textwidth}
        \centering
        \includegraphics[width=\linewidth]{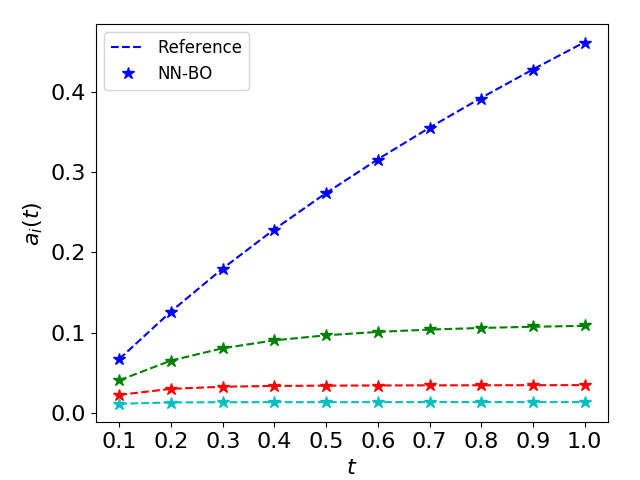}
        \caption{}\label{fig:nonlinear_inverse_ai}
    \end{subfigure}%
    \begin{subfigure}{.5\textwidth}
        \centering
        \includegraphics[width=\linewidth]{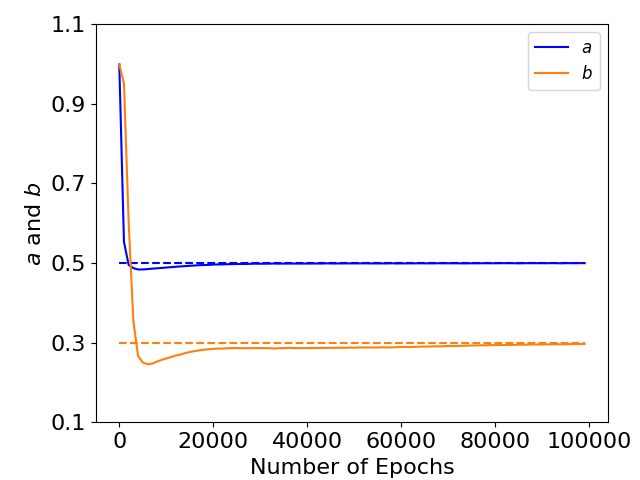}
        \caption{}\label{fig:nonlinear_inverse_a_b}
    \end{subfigure}
    \caption{Stochastic diffusion-reaction equation (inverse). Left: Evolution of scaling factors $a_i$ by NN-BO compared with the reference $a_i$ calculated using the standard numerical BO method for a forward problem; Right: Convergence of predicted $a$ and $b$ to the true hidden values during the training process.}
\end{figure}

Figure~\ref{fig:nonlinear_inverse_mean} and Figure~\ref{fig:nonlinear_inverse_var} shows the predicted solution mean and variance, respectively. Figure~\ref{fig:nonlinear_inverse_ui} and Figure~\ref{fig:nonlinear_inverse_ai} show the predicted BO modes $u_i$ and the scaling factors $a_i$. Table~\ref{tab:inverse_yerror} displays the root mean squared errors of the random coefficients $Y_i$. It is evident that when compared to the reference solutions, the NN-BO method is still accurate at solving the inverse problem. Finally, we display the convergence history of the predicted $a$ and $b$ in Figure~\ref{fig:nonlinear_inverse_a_b}, and we can observe that the inferred values converge to the true values after less than 100000 training epochs.

\section{Summary}\label{S:6}
To summarize, in this paper we presented two methods for solving time-dependent stochastic partial differential equations (SPDEs), i.e. the NN-DO method and the NN-BO method. They both make use of the expressiveness of Physics-Informed Neural Networks (PINNs). Similar to the standard dynamically orthogonal (DO) and bi-orthogonal (BO) methods, the proposed methods use either dynamical constraints on the spatial bases (NN-DO), or static constraints on both the spatial and the stochastic bases (NN-BO) to remove the time redundancy of the generalized Karhunen-L\`oeve expansion. Since the loss functions of neural networks can be directly established from an implicit form of the DO/BO constraints, the proposed methods are free from the assumptions needed for deriving the standard DO and BO equations, and thus they can be applied to a broader range of UQ problems. We demonstrated the performance of the NN-DO/BO methods with two artificially designed benchmark cases where exact DO/BO solutions can be derived, and we applied the NN-BO method to solve a time-dependent nonlinear diffusion reaction equation. Our numerical results show that the proposed NN-DO/BO methods are accurate for SPDEs with deterministic initial conditions and frequent eigenvalue crossings, and are reliable for long-time integration and high-dimensional random input. Moreover, additional flexibility over the standard BO/DO methods was demonstrated of the proposed methods in solving SPDEs by making direct use of the noisy scattered measurement data. They can seamlessly solve the time-dependent stochastic inverse problems by encoding the extra information into the loss function while tuning the hidden parameters at the training stage. These advantages were demonstrated in the last numerical example, and they exhibit the true potential of the NN-DO/BO method when applied to real physics/engineering applications.

However, there are two main current limitations of the NN-DO/BO methods and of PINNs in general. The first limitation is related to limited accuracy, i.e., the absolute errors cannot reach levels below about $10^{-5}$, due to the inherent inaccuracy of solving a non-convex optimization problem with no theoretical guarantees of a global minimum. Another limitation is the excessive cost associated with training the NN-DO/BO methods, especially for long-time integration. To this end, a promising approach is the use of parallel algorithms in time, such as the parareal algorithm~\cite{Lions2001}. For example, in the Burgers' equation example we could train all ten time-subdomains simultaneously and use the parareal algorithm iteration to obtain continuous in time solutions. This will be particularly effective if we use a lot of time-subdomains that can be trained in parallel. In fact, our preliminary experiments suggest that PINN training can be greatly accelerated using this approach for time-dependent PDEs, and this concept can also be extended to domain decomposition in space as well.

\section*{Acknowledgement}
This work is supported by ARL-Utah No.100028801-BROWN-APP (a sub-award of Cooperative Agreement W911NF-12-2-0023), NSF of China (No. 11671265) and the Science Challenge Project (No. TZ2018001).

\section*{References}
\bibliography{reference}

\end{document}